\documentclass{article}

\usepackage{graphicx}
\usepackage{subfigure} 
\usepackage[justification=centerfirst,singlelinecheck=false]{caption}
\usepackage{float}
\usepackage[section]{placeins}

\usepackage{natbib}

\usepackage{algorithm}
\usepackage{algorithmic}

\usepackage{amsmath}
\usepackage{amssymb}
\usepackage{comment}
\usepackage{bm}
\providecommand{\sign}[1]{\textrm{sign}\left(#1\right)}

\newcommand{\argmin}{\textrm{argmin}}

\usepackage{pgfplots}
\pgfplotsset{compat=newest}
\usepgfplotslibrary{groupplots}
\usepackage{pgfplotstable}
\usepackage{booktabs}
\usepackage{colortbl}
\usepackage{verbatim}
\usepackage{multirow}
\usepackage{picinpar}
\usepackage{fmtcount}
\usepackage[hmargin=1in,vmargin=1.5in]{geometry}

\usepackage[accepted]{icml2012} 

\icmltitlerunning{Non-convex polynomial loss functions for quantum annealing}

\begin{document}
\twocolumn[
\icmltitle{Construction of non-convex polynomial loss functions for training a binary classifier with quantum annealing}

\icmlauthor{Ryan Babbush}{babbush@fas.harvard.edu}
\icmlauthor{Vasil Denchev}{denchev@google.com}
\icmlauthor{Nan Ding}{dingnan@google.com}
\icmlauthor{Sergei Isakov}{iserge@google.com}
\icmlauthor{Hartmut Neven}{neven@google.com}
\icmladdress{Google,  340 Main St, Venice, CA 90291}
\icmlkeywords{binary classification, non-convex loss functions, quantum computing}

\vskip 0.3in
]

\begin{abstract}
Quantum annealing is a heuristic quantum algorithm which exploits quantum resources to minimize an objective function embedded as the energy levels of a programmable physical system. To take advantage of a potential quantum advantage, one needs to be able to map the problem of interest to the native hardware with reasonably low overhead. Because experimental considerations constrain our objective function to take the form of a low degree PUBO (\textsc{polynomial unconstrained binary optimization}), we employ non-convex loss functions which are polynomial functions of the margin. We show that these loss functions are robust to label noise and provide a clear advantage over convex methods. These loss functions may also be useful for classical approaches as they compile to regularized risk expressions which can be evaluated in constant time with respect to the number of training examples.
\end{abstract}

\section{Introduction}
\subsection{Quantum annealing}

While it is well known that gate model quantum algorithms provide an exponential speedup over the best known classical approaches for some problems \cite{shor1997,kitaev2002}, we are still technologically far from the ability to construct a large scale quantum computer which can robustly implement such algorithms for nontrivial problem instances. By contrast, rapid advances in superconducting qubit technology \cite{martinis2014} have provided a scalable platform for engineering medium-scale, controllable quantum systems at finite temperature. Such devices would be able to implement a quantum version of simulated annealing \cite{kirkpatrick1983} known as quantum annealing \cite{nishimori1998,farhi2000,santoro2002,somma2008}.

Because it is \textsc{NP-Hard} to determine the lowest energy configuration of a system of binary spins subject to controllable linear and quadratic energy terms \cite{barahona1982}, the ability to engineer and cool such a system provides an approach to solving any optimization problem in the class \textsc{NP}. In general, we do not expect that any device can efficiently solve instances of \textsc{NP-Hard} problems in the worst case. However, there is evidence that quantum resources such as tunneling and entanglement are generic computational resources which may help to solve problem instances which would be otherwise intractable for classical solvers. For instance, quantum annealing allows disordered magnets to relax to states of higher magnetic susceptibility asymptotically faster than classical annealing \cite{aeppli1999} and can solve certain oracular problems exponentially faster than any classical algorithm \cite{somma2012}. 

For the last few years, \emph{D-Wave Systems} has been commercially manufacturing quantum annealing machines \cite{Johnson2011}. These machines are the subject of ongoing scientific investigations by several third parties which aim to characterize the extent to which the hardware utilizes quantum resources and whether a scaling advantage is apparent for any class of problems \cite{boixo2014}.

\subsection{Training under non-convex loss}

The problem we consider in this work is the training of a linear binary classifier using noisy data \cite{bishop2006}. We assume that the training data is provided as a matrix $\bm{\hat{x}} \in \mathbb{R}^{m \times n}$ with the $m$ rows corresponding to unique descriptor vectors containing $n$ features. We are also provided with a vector of labels, $\bm{y} \in \left\{-1, 1\right\}^m$, which associate a binary classification with each feature vector. The training problem is to determine an optimal classifier $\bm{w} \in \mathbb{R}^n$ which predicts the data by classifying example $i$ as $\sign{\bm{w}^\top \bm{x}_i}$.

The classifier may be viewed as a hyperplane in feature space which divides data points into negative and positive classifications. In this space, the distance that example $i$ falls from the classification hyperplane $\bm{w}$ is referred to as the margin $\gamma_i  \equiv y_i \,\bm{x}_i^\top \bm{w}$. Whereas a negative margin represents a classification opposite the training label, a positive margin represents a classification consistent with the training label. To cast training as an optimization problem we use the concept of a loss function which penalizes the classification of each example according to its margin \cite{bishop2006}. Perhaps the simplest loss function is the 0-1 loss function which provides a correct classification with penalty 0 and an incorrect classification with penalty 1,
\begin{equation}
L_{01}\left(\gamma_i\right) \equiv \frac{1-\sign{\gamma_i}}{2}.
\end{equation}
The training objective (known in machine learning as total empirical risk) is given as the mean loss over all examples in the training set. For instance, the 0-1 empirical risk function is
\begin{equation}
f_{01}\left(\bm{w}\right) \equiv \frac{1}{m}\sum_{i = 0}^{m-1} L_{01}\left(\gamma_i\right).
\end{equation}

Unfortunately, minimization of the 0-1 empirical risk function is known to be \textsc{NP-Hard} \cite{feldman2012}. For this reason, most contemporary research focuses on convex loss functions which are provably efficient to optimize. However, in data with high label noise, this is an unacceptable compromise as the efficiency gained by convex minimization allows only for the efficient computation of a poor classifier \cite{manwani2013}. By contrast, training under non-convex loss functions is known to provide robust classifiers even when nearly half of the examples are mislabeled \cite{long2010}.

Objectives such as these, for which certain instances may require exponential time using classical heuristics, are ideal candidates for quantum annealing. In order to attempt non-convex risk minimization with quantum annealing in the near future, one must first \emph{efficiently} compile the problem to a form compatible with quantum hardware. Due to engineering considerations, this usually means preparing the problem as an instance of QUBO (\textsc{quadratic unconstrained binary optimization}). Previously, Denchev \emph{at al.} introduced a method for mapping non-convex loss training to QUBO \cite{denchev2012} for the purposes of solving on a quantum device. However, in that work, the number of variables required to accomplish the embedding was lower-bounded by the number of training examples. While clearly robust, this scheme seems impractical for medium-scale quantum annealers due to the large qubit overhead.

Here, we develop a different embedding in which the number of required variables is independent of the number of training examples. This is accomplished by deriving loss functions which are polynomial functions of the margins. We show that such loss functions give rise to empirical risk objectives expressible as PUBO. Compatibility with quantum hardware comes from the fact that any PUBO can be reduced to QUBO using a number of boolean ancilla variables that is at most $O\left(N^{2 \log k}\right)$ where $N$ is the number of logical variables and $k$ is the order of the PUBO \cite{boros2012}. Coincidentally, this implies that the empirical risk objective associated with any polynomial loss function can be evaluated in an amount of time that does not depend on the number of training examples.

In particular, we investigate the use of third-order and sixth-order polynomial loss functions. The cubic loss function is chosen as $k=3$ is the lowest order that gives us non-convexity. Polynomial loss has very different characteristics depending on the parity of $k$ so we also investigate an even degree polynomial loss function. We forgo quartic loss in favor of sixth-order loss as the latter qualitatively fits 0-1 loss much better than the former. After deriving optimal forms of cubic loss and sixth-order loss we numerically investigate the properties of these loss functions to show robustness to label noise. Finally, we demonstrate an explicit mapping of any polynomial risk objective to a tensor representing an instance of PUBO that is easily compiled to quantum hardware.

\section{Cubic loss}

In this section we derive an approximate embedding of 0-1 risk under $\ell_2$-norm regularization as a cubic function of the weights. We begin by considering the general forms of the cubic loss and cubic risk functions,
\begin{eqnarray}
L_3 \left(\gamma_i\right) & = & \alpha_0 + \alpha_1 \gamma_i + \alpha_2 \gamma_i^2 + \alpha_3 \gamma_i^3\\
f_3\left(\bm{w}\right) & = & \frac{1}{m}\sum_{i=0}^{m-1} L_3 \left(\gamma_i\right).
\end{eqnarray}
Thus, the embedding problem is to choose the optimal $\bm{\alpha} \in \mathbb{R}^4$ so that $f_3\left(\bm{w}\right)$ best approximates $f_{01}\left(\bm{w}\right)$. To accomplish this we consider the $\ell_2$-norm between 0-1 risk and cubic risk,
\begin{equation}
\bm{\alpha}^* = \argmin\left\{\int P\left(\bm{w}\right) \left[f_{01}\left(\bm{w}\right) - f_3\left(\bm{w}\right)\right]^2 \textrm{d} \bm{w}\right\}.
\end{equation}
Here, $P\left(\bm{w}\right)$ is the prior distribution of the weights. If we incorporate an $\ell_2$-norm regularizer, $\Omega_2\left(\bm{w}\right)$, into our ultimate training objective, $E\left(\bm{w}\right)$, i.e.
\begin{eqnarray}
\Omega_2\left(\bm{w}\right) & = & \frac{\lambda_2}{2} \,\bm{w}^\top \bm{w}\\
E\left(\bm{w}\right) & = & f\left(\bm{w}\right) + \Omega_2\left(\bm{w}\right),
\end{eqnarray}
then we are provided with a Gaussian prior on the weights \cite{rennie2003} taking the form,
\begin{equation}
P\left(w_i\right) = \sqrt{\frac{\lambda_2}{2 \pi}} e^{- \lambda_2 w_i^2 / 2}.
\end{equation}
Immediately, we see that for the optimal solution $\alpha_2 \rightarrow 0$ since 0-1 loss is an odd function and the least squares residual is weighed over a symmetric function (the Gaussian prior). Furthermore, we can ignore $\alpha_0$ and the constant factor of $\frac{1}{2}$ in $L_{01}\left(\gamma_i\right)$ as these constants are irrelevant for the training problem. With this in mind, we expand the empirical risk functions under the integral in the embedding problem as,
\begin{equation}
\int P\left(\bm{w}\right) \left(\sum_{i=0}^{m-1} \frac{\sign{\gamma_i}}{2} + \alpha_1\gamma_i + \alpha_3\gamma_i^3\right)^2 \textrm{d}\bm{w}.
\end{equation}
Thus,
\begin{equation}
\bm{\alpha}^* = \argmin\left\{\sum_{i=0}^{m-1}\sum_{j=0}^{m-1} \int P\left(\bm{\gamma}\right) F_{ij} \left(\bm{\gamma}\right)\textrm{d}\bm{\gamma} \right\}
\end{equation}
where
\begin{align}
F_{ij}\left(\bm{\gamma}\right) \quad \equiv & \quad \frac{\alpha_1}{2}\left[\gamma_i \sign{\gamma_j} + \gamma_j \sign{\gamma_i}\right]\\
&  + \frac{\alpha_3}{2}\left[\gamma_i^3 \sign{\gamma_j} + \gamma_j^3 \sign{\gamma_i}\right]\nonumber\\
&  + \alpha_1^2 \gamma_i \gamma_j + \alpha_1 \alpha_3 \left(\gamma_i^3 \gamma_j + \gamma_j^3 \gamma_i\right) + \alpha_3^2 \gamma_i^3 \gamma_j^3 \nonumber.
\end{align}

Without loss of generality, we may assume that $P\left(\bm{\gamma}\right)$ is a multinormal distribution centered at zero with a covariance matrix,
\begin{equation}
\bm{\hat{\Sigma}} = \frac{1}{\lambda_2} \left(\bm{\hat{x}}^\top \bm{\hat{x}}\right) \odot \left(\bm{y} \, \bm{y}^\top\right)
\end{equation}
where $\odot$ implies element-wise matrix multiplication (i.e. the Hadamard product). The multinormal distribution occurs because the margins arise as the result of the training examples being projected by classifiers drawn from the prior distribution given by $P\left(\bm{w}\right)$. Since each weight is normally distributed with zero mean and variance $\lambda_2^{-2}$ in the prior, the distribution of margins associated with training example $i$ will be a Gaussian with zero mean and variance,
\begin{equation}
\sigma_i^2 = \frac{1}{n} \sum_{j=0}^{n-1} \left(\frac{x_{ij}}{\lambda_2}\right)^2.
\end{equation}
Because each linear combination of the elements of the margin vector is also normally distributed, we have a multinormal distribution. This is true regardless of the number of features or any particular qualities of the training data.

Accordingly, if we wished to scale $\bm{w}$ to a range which contains $\bm{w}^*$ with a likelihood in the $r^\textrm{th}$ standard deviation of the prior then we should make $\bm{w} \in \left[- \frac{r}{\sqrt{\lambda_2 m}},  \frac{r}{\sqrt{\lambda_2 m}}\right]^n$. However, making $r$ too large would be problematic because this could allow the cubic term to dominate the quadratic regularizer. This necessitates a cutoff on the maximum weight value to ensure that unbounded cubic losses associated with large negative margins do not overcome the regularizer. In practice, $r$ would need to be selected as a hyperparameter.

Since the integrand has only two point correlation functions, we can integrate over the marginal distribution of $\gamma_i$ and $\gamma_j$ which is a binormal distribution with covariance,
\begin{eqnarray}
\bm{\hat{\Sigma}}_{ij} & = & \frac{1}{\lambda_2} \left(\begin{matrix} \bm{x}_i^\top\! \bm{x}_i\,\, & y_i y_j\bm{x}_i^\top\! \bm{x}_j\\
y_j y_i\bm{x}_j^\top \!\bm{x}_i\,\, & \bm{x}_j^\top\!\bm{x}_j\end{matrix}\right)\\
& = & \left(\begin{matrix} \sigma_i^2 & \rho_{ij}\,\sigma_i \sigma_j \\
\rho_{ij} \,\sigma_j \sigma_i & \sigma_j^2 \end{matrix}\right).
\end{eqnarray}
We can analytically evaluate the double integral,
\begin{align}
\bm{\alpha}^* = & \,\, \argmin\left\{\sum_{i=0}^{m-1}\sum_{j=0}^{m-1} I_{ij}\right\}\\
I_{ij} = & \int_{-\infty}^{\infty} \int_{-\infty}^{\infty} P_{ij} \left(\bm{\gamma}\right) F_{ij} \left(\bm{\gamma}\right) \textrm{d}\gamma_i \textrm{d}\gamma_j\\
 = & \underbrace{\frac{\rho_{ij}\left(\sigma_i + \sigma_j\right)}{\sqrt{2 \pi}}}_{t_0} \alpha_1
+ \underbrace{\frac{\rho_{ij} \left(3-\rho_{ij}^2\right)\left(\sigma_i^3 +\sigma_j^3\right)}{\sqrt{2 \pi}}}_{t_1} \alpha_3 \nonumber \\
& +\underbrace{3\,\rho_{ij} \sigma_i \sigma_j \left(\sigma_i^2 + \sigma_j^2\right)}_{t_2}  \alpha_1 \alpha_3 
 +  \underbrace{\rho_{ij}\sigma_i \sigma_j}_{t_3}\alpha_1^2\nonumber\\
& + \underbrace{3\,\rho_{ij}\left(3 + 2 \rho_{ij}^2\right)\sigma_i^3 \sigma_j^3}_{t_4} \alpha_3^2 \nonumber\\
 = & \,\,t_0 \alpha_1 + t_1 \alpha_3 + t_2 \alpha_1 \alpha_3 + t_3 \alpha_1^2 + t_4 \alpha_3^2\nonumber
\end{align}

With the integral in closed form, we obtain the argmin using simple algebra by solving,
\begin{equation}
\nabla \left(\sum_{i = 0}^{m-1}\sum_{j = 0}^{m-1} I_{ij} \right) = 0.
\end{equation}
The analytical coefficients for a single set of training examples are,
\begin{equation}
\alpha_1^* = \frac{2\, t_0 t_4 - t_1 t_2}{t_2^2 - 4\, t_3 t_4}\quad\quad\quad \alpha_3^* = \frac{2\, t_1 t_3 - t_0 t_2}{t_2^2 - 4\, t_3 t_4}.
\end{equation}

Thus, for the full training set we must sum together the $t$ values from each set of $\left(i,j\right)$ before plugging values into the expression for $\alpha_1$ and $\alpha_3$. We note that the computation required to obtain these coefficients is $O\left(m^2\right)$. Figure 1 shows several cubic loss function fits associated with various real data sets from the UCI Machine Learning Repository.

The coefficients above are analytic and optimal for embedding 0-1 risk. However, it is instructive to explain what would happen if we had chosen to fit the loss function instead of the objective function. This would have produced a far simpler embedding problem\footnote{Note the difference between $\bm{\alpha}^\star$ and $\bm{\alpha}^*$.},
\begin{equation}
\bm{\alpha}^\star = \argmin\left\{\int_{-\infty}^{\infty} P^\star \left(\gamma\right) \left[L_{01}\left(\gamma\right) - L_3\left(\gamma\right)\right]^2 \textrm{d}\gamma\right\}
\end{equation}
where
\begin{equation}
P^\star\left(\gamma\right) \equiv \frac{1}{\sigma \sqrt{2\pi}} e^{-\gamma^2 / 2\sigma^2}, \,\, \sigma^2 = \frac{1}{ \lambda_2 m}\textrm{tr}\left[\bm{\hat{x}}^\top \bm{\hat{x}}\right].
\end{equation}
This time the integral is trivial to evaluate,
\begin{eqnarray}
I^\star & = & \int_{-\infty}^{\infty} G^\star\left(\gamma,\right) \left[L_{01}\left(\gamma\right) - L_3\left(\gamma\right)\right]^2 \textrm{d}\gamma\\
& = & \sqrt{\frac{2}{\pi}} \sigma \alpha_1 + \sigma^2 \alpha_1^2 + 2 \sqrt{\frac{2}{\pi}}\sigma^3 \alpha_3\nonumber\\
& & + \,\, 6\,\sigma^4 \alpha_1 \alpha_3 + 15\, \sigma^6 \alpha_3^2 \nonumber.
\end{eqnarray}
As before, convexity guarantees that $\nabla I$ will have exactly one real root which we find analytically,
\begin{equation}
\alpha_1^\star = -\frac{3}{2 \sqrt{2\pi} \sigma}\quad \quad \quad \alpha_3^\star = \frac{1}{6 \sqrt{2 \pi} \sigma^3}.
\end{equation}
These result seems much simpler than when we embed the entire risk function but they are obviously less useful. However, if we make the assumption that $\sigma_i = \sigma_j = \sigma$ then, $\alpha_1^\star = \alpha_1^*$ and $\alpha_3^\star = \alpha_3^*$.
\begin{figure}[ht]
		\includegraphics[scale = .4]{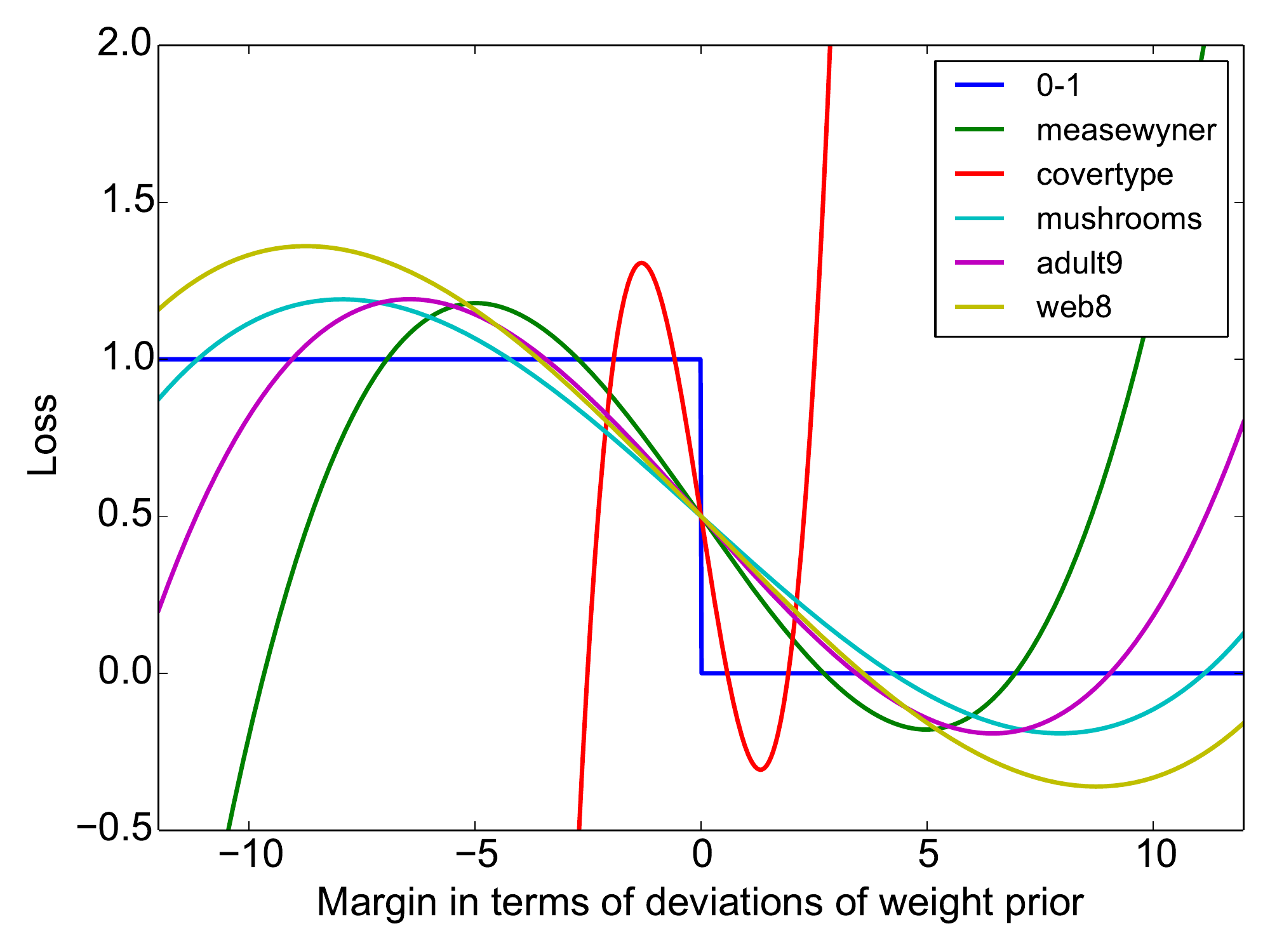}
\vspace{-.25cm}
\caption{Cubic loss fits for a variety of real data sets. Due to the different properties of their correlation matrices each set is associated with unique cubic loss coefficients. \label{loss_functions}}
\end{figure}

In Figure 2 we study the performance of our embedded loss function by exactly enumerating the solution space produced by small synthetic data sets. These data sets were produced by randomly generating classifiers with weights drawn from a normal distribution and then using that classifier to label feature vectors with features drawn from a uniform distribution. Symmetric label errors are then manually injected at random. A maximum weight cutoff is imposed at the second standard deviation of the weight prior.  Further numerical analysis of cubic loss on synthetic data sets is included in the Appendix.

While the cubic loss embedding is somewhat noisy in the sense that it does not perfectly approximate 0-1 loss, it is clearly robust in the sense that test error does not depend strongly on label error for up to 45\% label noise. This remains true whether we consider the best fifty states embedded in cubic loss or only the absolute ground state. These results indicate that cubic loss has an advantage over convex methods when data is known to contain substantial label noise.

\section{Sixth-order loss}

One potentially unattractive feature of the cubic loss function is that it is necessary to fix the scale of the weights as a hyperparameter. Since we intend to encode our objective function as QUBO for quantum annealing, we will need to choose a maximum weight. While one can prove that the optimal classifier will have weights in the interval $\left[-\frac{1}{\sqrt{\lambda_2}}, \frac{1}{\sqrt{\lambda_2}}\right]$, such a large range is potentially problematic for regularized cubic risk as the loss associated with large negative margins tends towards negative infinity faster than the $\ell_2$ regularizer can penalize the large weights which would produce such margins.
\begin{figure*}[ht]
	\centering
	\begin{subfigure}{}
		\includegraphics[scale = .4]{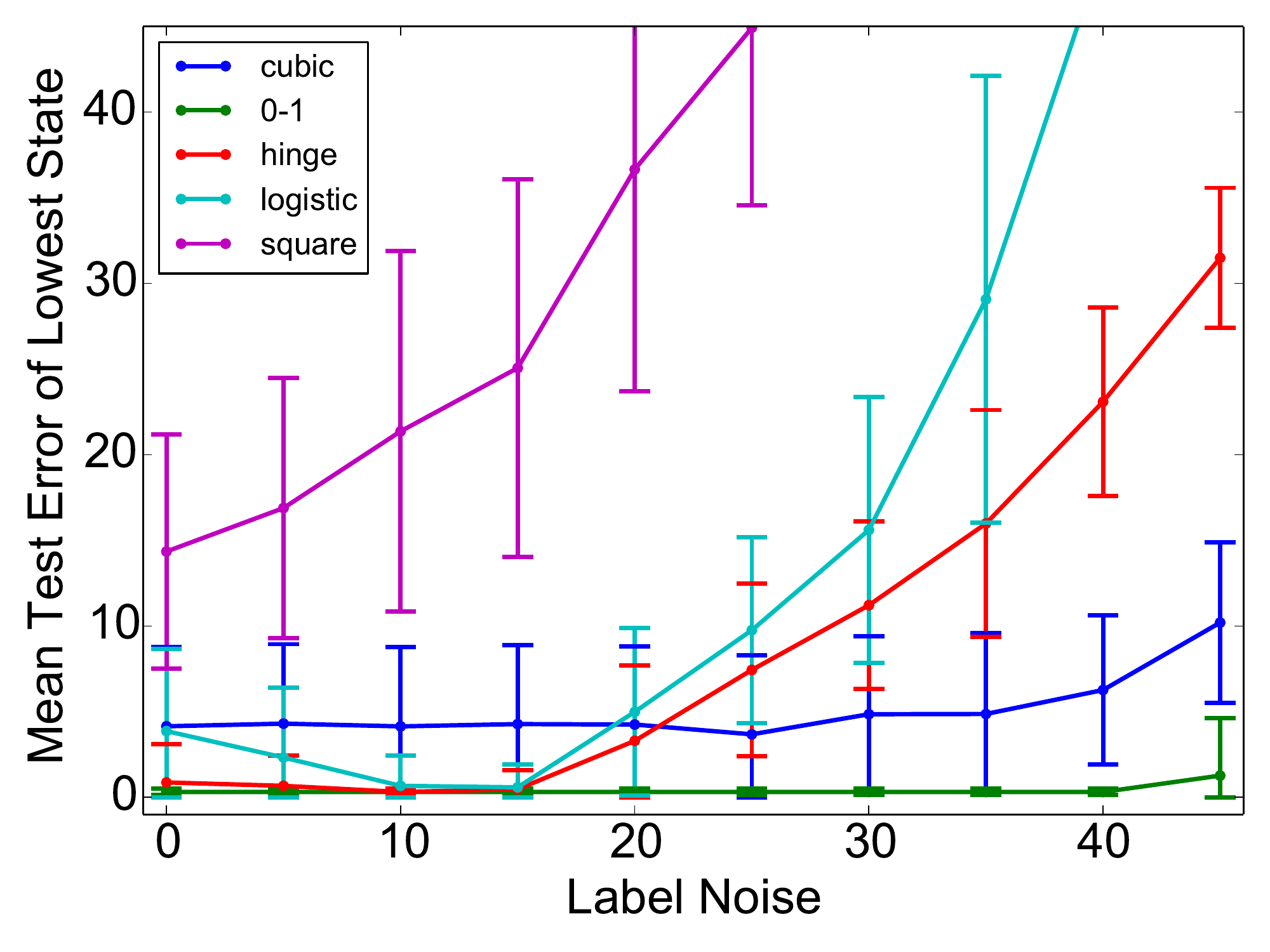}
	\end{subfigure}
	\begin{subfigure}{}
		\includegraphics[scale = .4]{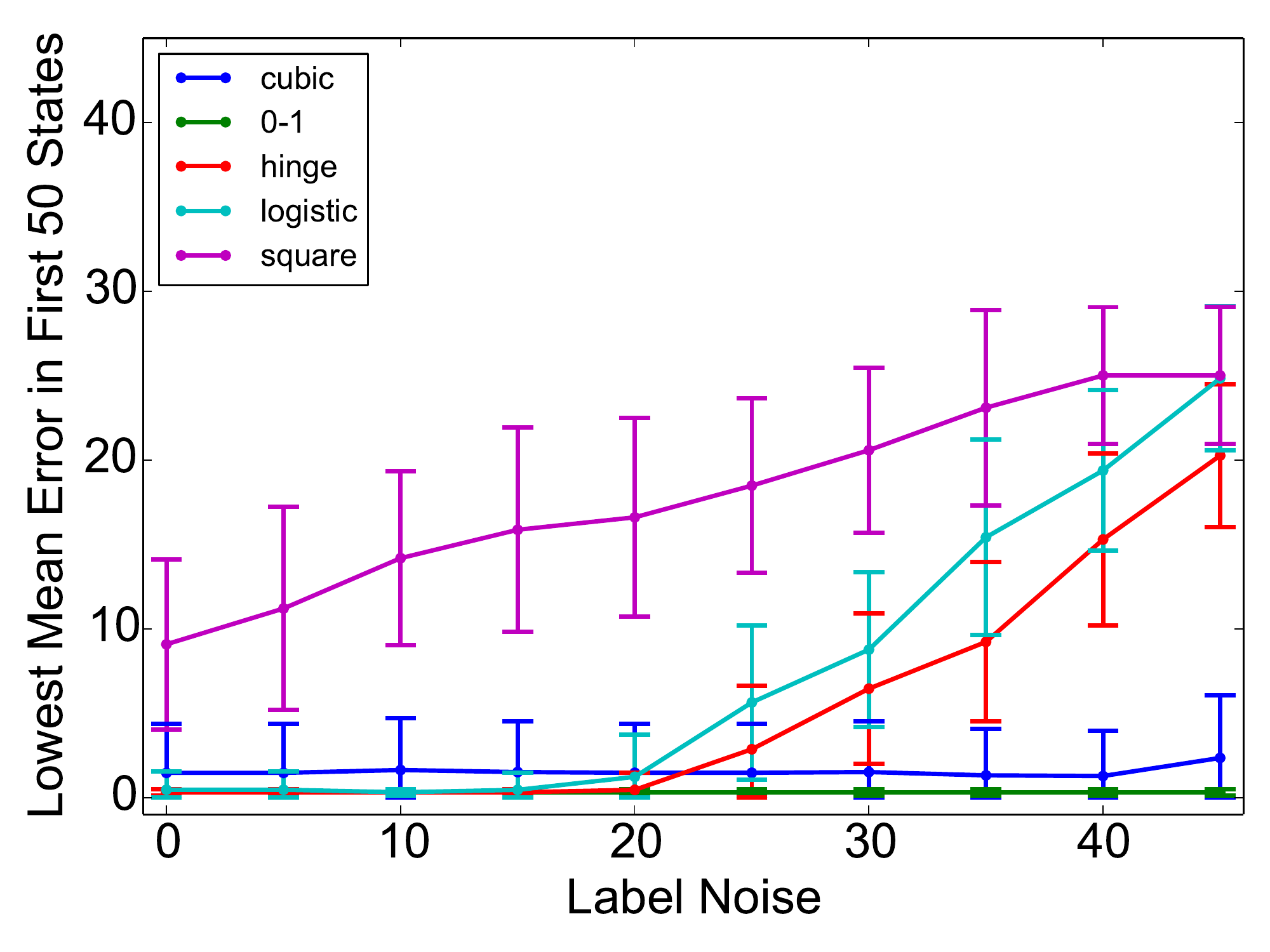}
	\end{subfigure}
	\vspace{-.25cm}
	\begin{subfigure}{}
		\includegraphics[scale = .4]{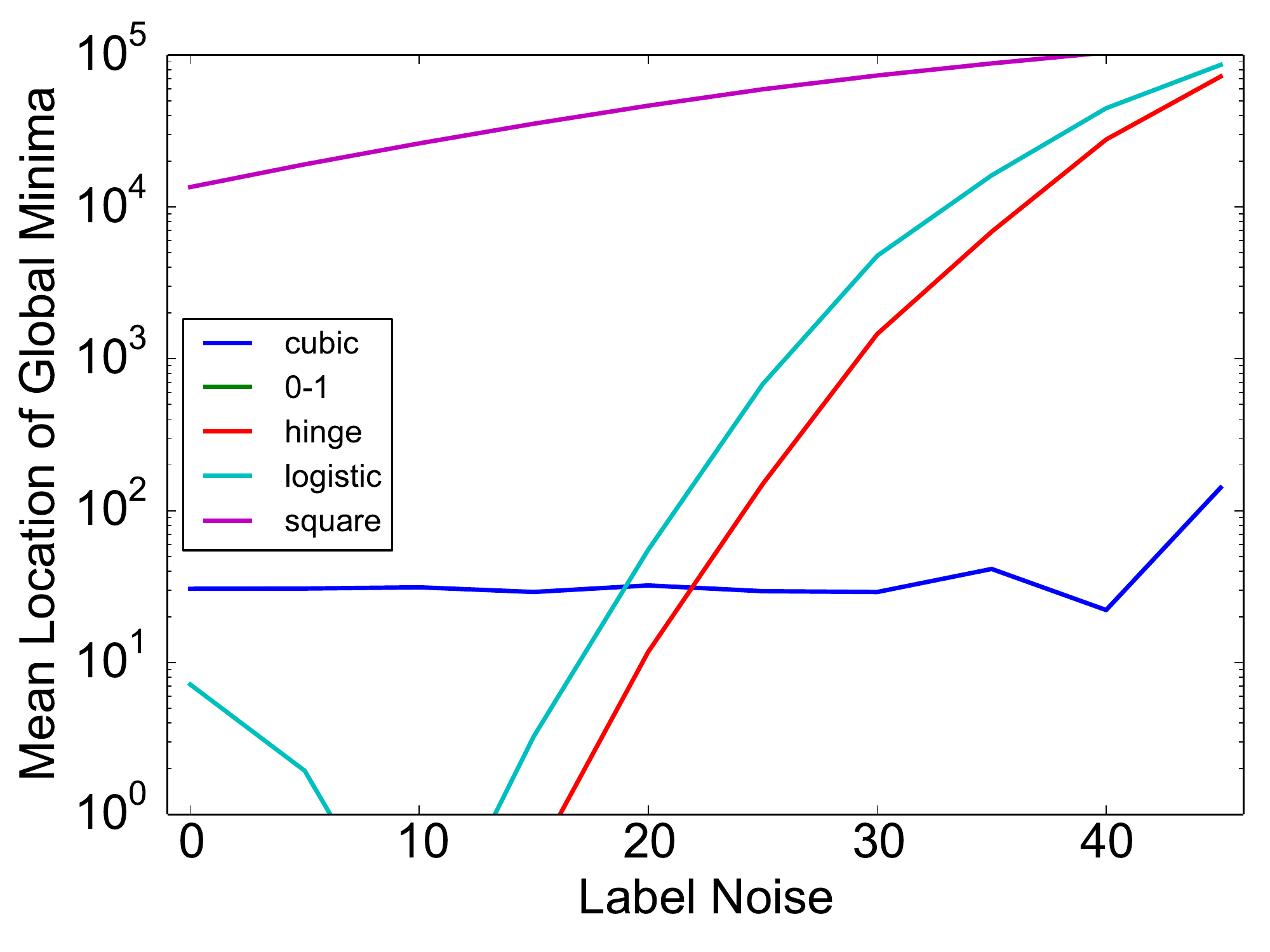}
	\end{subfigure}
	\begin{subfigure}{}
		\includegraphics[scale = .4]{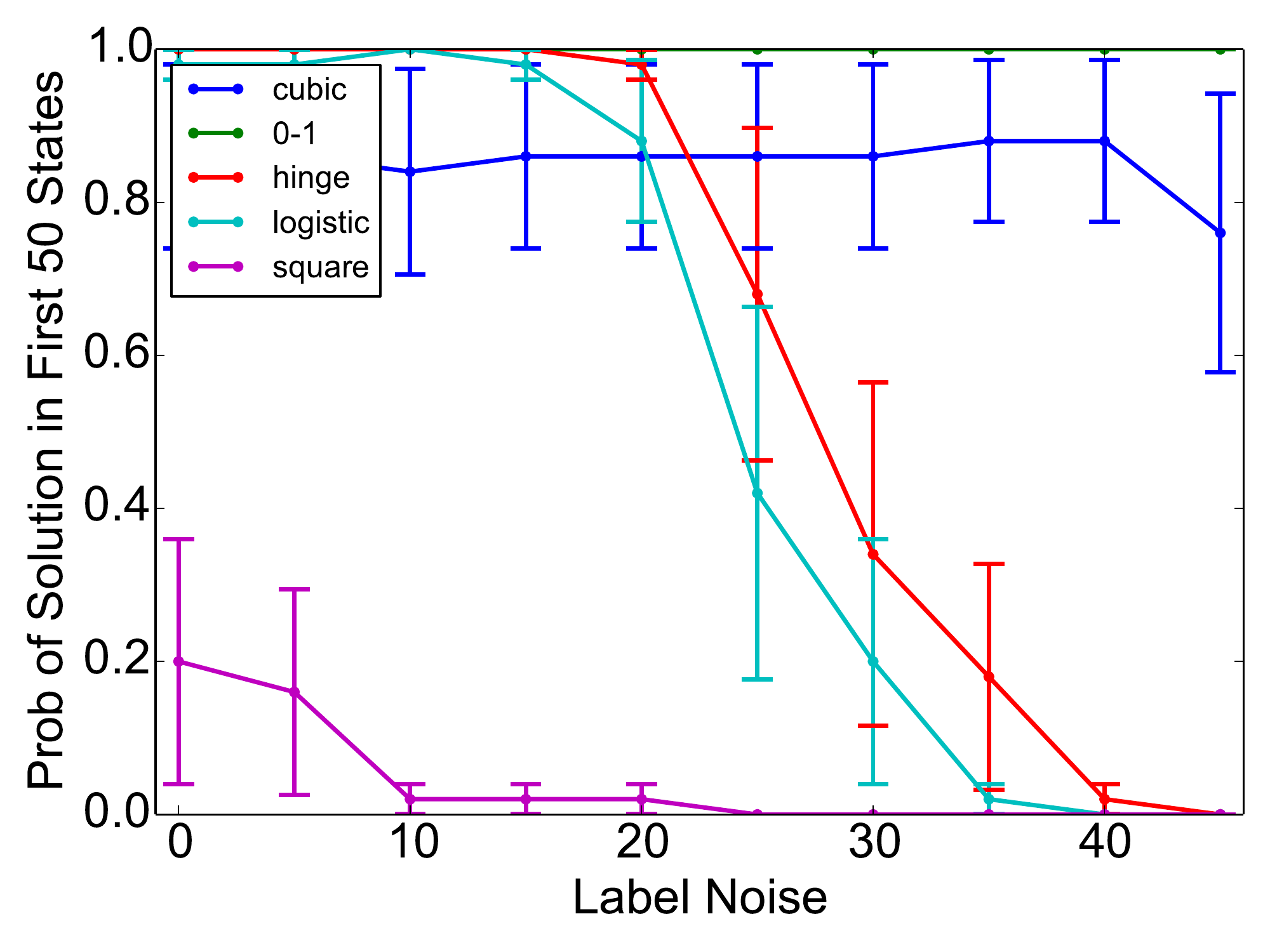}
	\end{subfigure}
	\caption{Performance of cubic loss on synthetic data sets with $10^4$ examples, 9 features and a bit depth of 2. We exactly enumerate all fixed bit depth classifiers and evaluate the empirical risk under various loss functions. Error bars are obtained by repeating the experiment on fifty data sets. The upper left plot shows the error in the lowest objective state and the upper right plot shows the error in the best state of the lowest fifty. The lower left plot shows mean position of the global minima in the eigenspectra of the various objective functions. The lower right plot shows the probability of the global minima being in the first fifty states. As we can see, the global minima remains very near the bottom of the eigenspectra for cubic loss, regardless of label noise.\label{test_error}}
\vspace{.25cm}
\end{figure*}

Alternatively, one might consider using a polynomial loss function of even degree as such loss functions will not diverge to negative infinity for large negative margins. In order to do this, we must fix the highest order term in the loss function as a hyperparameter. This is because 0-1 loss is an odd function so if we attempt to embed 0-1 risk in an even degree polynomial, the even terms will vanish. Accordingly, we turn our attention to the sixth-order loss and empirical risk functions,
\begin{eqnarray}
L_6\left(\gamma_i\right) & \equiv & \omega \gamma_i^6 + \sum_{k=1}^{5} \beta_k \gamma_i^k\\
f_6\left(\bm{w}\right) & \equiv & \frac{1}{m}\sum_{i=0}^{m-1} L_3 \left(\gamma_i\right).
\end{eqnarray}

Here, $\omega$ is taken to be a hyperparameter. We will solve for the values of $\bm{\beta}$. In the case of the cubic loss function, we used the weight prior imposed by $\ell_2$-norm regularization and data set covariance to derive a margin prior which was used for embedding. However, this is unnecessary for the sixth-order loss function as $\omega$ provides a very simple prior on the margins,
\begin{equation}
P\left(\gamma\right) = \frac{\omega^{1/6}}{2\, \Gamma\left(7/6\right)}e^{-\omega \gamma^6}
\end{equation} 
where $\Gamma$ is the standard gamma function. With this definition, the embedding problem becomes
\begin{equation}
\bm{\beta}^* = \argmin\left\{\int P\left(\gamma\right) \left[L_{01}\left(\gamma\right) - L_6\left(\gamma\right)\right]^2 \textrm{d} \gamma \right\}.\label{integral6}
\end{equation}

\begin{figure}[ht]
	\centering
	\includegraphics[scale = .4]{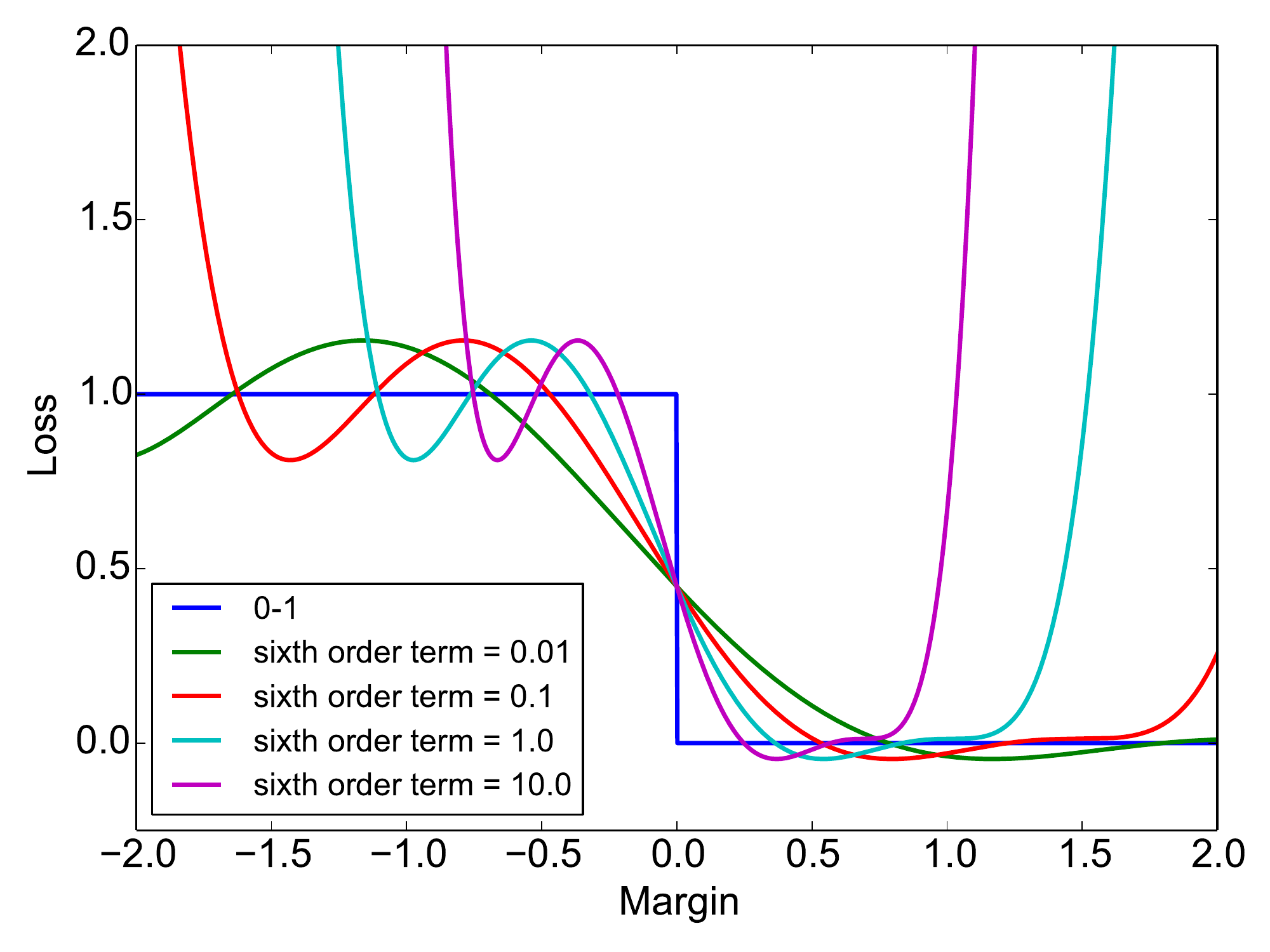}
	\vspace{-.25cm}
\caption{The sixth-order loss function at various values of the fixed sixth-order coefficient, $\omega$. This coefficient is taken to be a hyperparameter.}
\vspace{-.25cm}
\end{figure}
Whereas we chose $\bm{\alpha}$ for cubic loss by fitting the empirical risk function, we choose $\bm{\beta}$ for sixth-order loss by fitting the loss function directly. Since the sixth-order loss function is already parameterized in terms of a hyperparameter, there is nothing to gain by devising a more elaborate fit based on empirical risk. After evaluating $I_6$, the integral in Eq.~\ref{integral6}, we can obtain $\bm{\beta}^*$ by solving, $\nabla I_6 = 0$. The optimal values of $\bm{\beta}$ are included in the Appendix.  Figure 3 shows the sixth-order loss function for various values of $\omega$.

Figure 4 (on the next page) shows the performance of the sixth-order loss function on selected data sets from the UCI Machine Learning Repository. To stand-in for a quantum annealer, we optimized the sixth-order objective function using a simulated annealing routine which was run for over one hundred thousand CPU hours. In addition to standard convex methods, we compare to 0-1 loss optimized using the same simulated annealing code run with the same number of restarts and variable updates as were used to optimize sixth-order loss. We also include the ``q-loss'' results from \cite{denchev2012} which were obtained for that work using another metaheuristic algorithm (Tabu search). Details regarding the 10-fold cross validation procedure are reported in the Appendix.

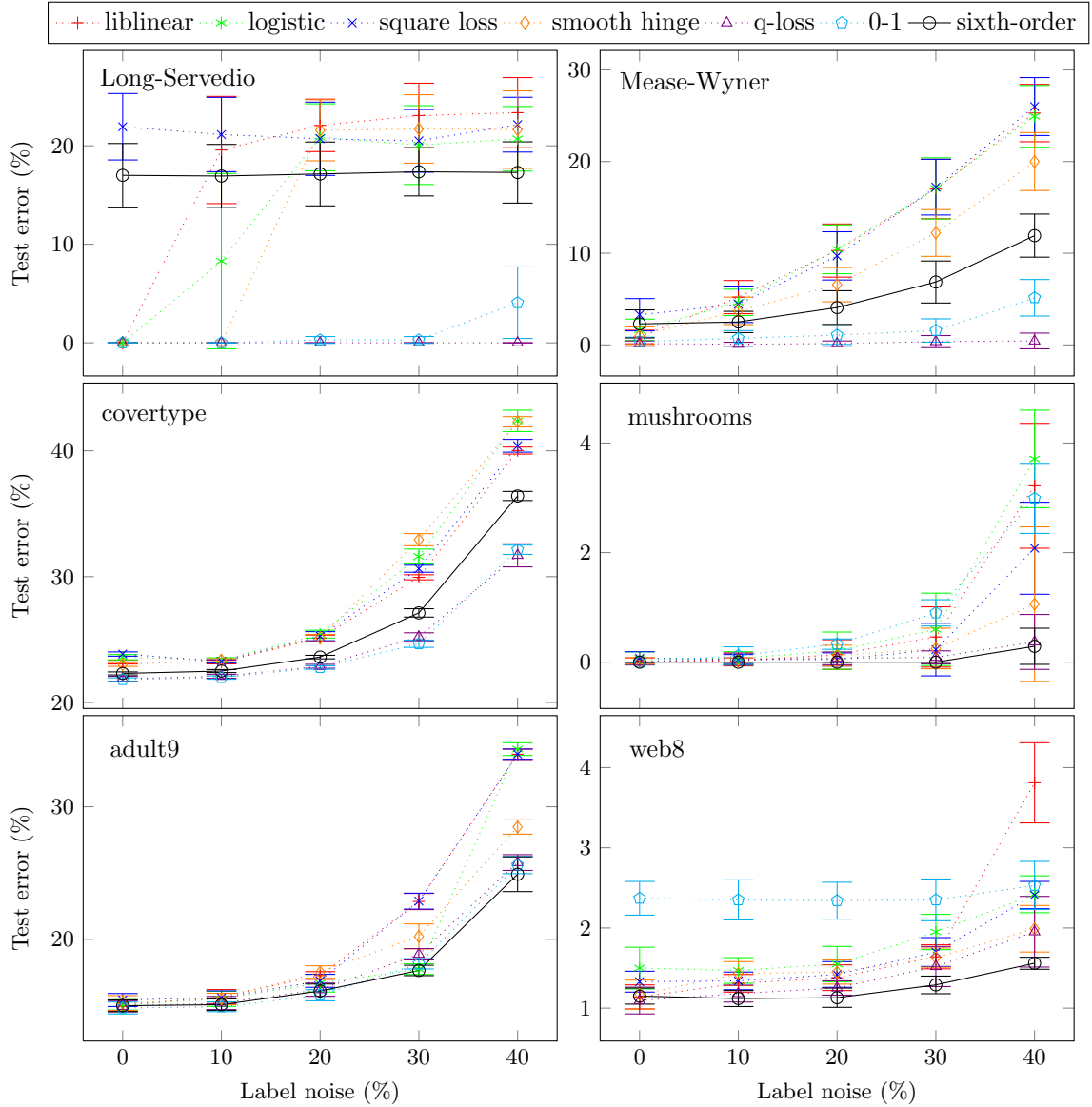
\begin{figure*}
\centering
\tikzset{every mark/.append style={solid,scale=1.2}}
\pgfplotsset{every axis/.append style={line width=0.4pt,xtick = {0,10,20,30,40},xticklabels={},height=6.13cm,width=8.2cm,cycle list ={
{red,dotted,mark=+},
{green,dotted,mark=asterisk},
{blue,dotted,mark=x},
{orange,dotted,mark=diamond},
{violet,dotted,mark=triangle},
{cyan,dotted,mark=pentagon},
{black,solid,mark=o},
}},
every error bar/.append style={solid}}
\begin{tikzpicture}                             
	\begin{axis} [name=plot1,legend columns=-1,legend entries={liblinear,logistic,square loss,smooth hinge, q-loss, 0-1, sixth-order},
	legend style={at={(-.075,1.02)}, anchor=south west},tick label style={font=\small},y label style={font=\small},ylabel=Test error ($\%$)]
	\node at (axis description cs:0.2,0.91) {Long-Servedio};

	\addplot+[error bars/.cd, y dir=both, y explicit, error mark=|,error mark options={scale=3.0}]
		coordinates {
			(0,0) +- (0,0)
			(10,19.57) +- (0,5.44)
			(20,22.07) +- (0,2.66)
			(30,23.07) +- (0,3.28)
			(40,23.36) +- (0,3.56)
		};

	\addplot+[error bars/.cd, y dir=both, y explicit, error mark=|,error mark options={scale=3.0}]
		coordinates {
			(0,0) +- (0,0)
			(10,8.29) +- (0,8.9)
			(20,20.86) +- (0,3.4)
			(30,20.07) +- (0,4)
			(40,20.71) +- (0,3.28)
		};

	\addplot+[error bars/.cd, y dir=both, y explicit, error mark=|,error mark options={scale=3.0}]
		coordinates {
			(0,21.93) +- (0,3.37)
			(10,21.14) +- (0,3.77)
			(20,20.71) +- (0,3.72)
			(30,20.5) +- (0,3.18)
			(40,22.14) +- (0,2.78)
		};

	\addplot+[error bars/.cd, y dir=both, y explicit, error mark=|,error mark options={scale=3.0}]
		coordinates {
			(0,0) +- (0,0)
			(10,0) +- (0,0)
			(20,21.57) +- (0,3.1)
			(30,21.71) +- (0,3.47)
			(40,21.64) +- (0,3.93)
		};
	\addplot+[error bars/.cd, y dir=both, y explicit, error mark=|,error mark options={scale=3.0}]
		coordinates {
(0,0.000000)+-(0,0.000000)
(10,0.000000)+-(0,0.000000)
(20,0.000000)+-(0,0.000000)
(30,0.000000)+-(0,0.000000)
(40,0.000000)+-(0,0.000000)
		};

	\addplot+[error bars/.cd, y dir=both, y explicit, error mark=|,error mark options={scale=3.0}]
		coordinates {
			(0,0) +- (0,0)
			(10,0) +- (0,0)
			(20,0.29) +- (0,.35)
			(30,0.29) +- (0,.35)
			(40,4.07) +- (0,3.63)
		};

	\addplot+[error bars/.cd, y dir=both, y explicit, error mark=|,error mark options={scale=3.0}]
		coordinates {
			(0,17) +- (0,3.23)
			(10,16.93) +- (0,3.21)
			(20,17.14) +- (0,3.24)
			(30,17.36) +- (0,2.45)
			(40,17.29) +- (0,3.11)
		};

	\end{axis}
	\begin{axis} [name=plot2,at={($(plot1.east)+(0.6cm,0)$)},anchor=west,tick label style={font=\small}]
	\node at (axis description cs:0.2,0.9) {Mease-Wyner};

	\addplot+[error bars/.cd, y dir=both, y explicit, error mark=|,error mark options={scale=3.0}]
		coordinates {
			(0,0.86) +- (0,0.74)
			(10,5.21) +- (0,1.81)
			(20,10.29) +- (0,2.9)
			(30,17) +- (0,3.24)
			(40,25.29) +- (0,3.13)
		};

	\addplot+[error bars/.cd, y dir=both, y explicit, error mark=|,error mark options={scale=3.0}]
		coordinates {
			(0,1.64) +- (0,1.17)
			(10,4.64) +- (0,1.44)
			(20,10.43) +- (0,2.66)
			(30,17.07) +- (0,3.34)
			(40,24.93) +- (0,3.37)
		};

	\addplot+[error bars/.cd, y dir=both, y explicit, error mark=|,error mark options={scale=3.0}]
		coordinates {
			(0,3.29) +- (0,1.76)
			(10,4.43) +- (0,1.99)
			(20,9.71) +- (0,2.63)
			(30,17.21) +- (0,3.04)
			(40,26) +- (0,3.16)
		};

	\addplot+[error bars/.cd, y dir=both, y explicit, error mark=|,error mark options={scale=3.0}]
		coordinates {
			(0,1) +- (0,0.96)
			(10,3.71) +- (0,1.5)
			(20,6.57) +- (0,1.87)
			(30,12.21) +- (0,2.55)
			(40,20) +- (0,3.16)
		};
	\addplot+[error bars/.cd, y dir=both, y explicit, error mark=|,error mark options={scale=3.0}]
		coordinates {
(0,0.142900)+-(0,0.285700)
(10,0.071400)+-(0,0.214300)
(20,0.142900)+-(0,0.285700)
(30,0.357100)+-(0,0.658500)
(40,0.428600)+-(0,0.857100)
		};

	\addplot+[error bars/.cd, y dir=both, y explicit, error mark=|,error mark options={scale=3.0}]
		coordinates {
(0,0.36)+-(0,.48)
(10,.71)+-(0,.85)
(20,1.07)+-(0,1.02)
(30,1.57)+-(0,1.27)
(40,5.14)+-(0,1.99)
		};
	\addplot+[error bars/.cd, y dir=both, y explicit, error mark=|,error mark options={scale=3.0}]
		coordinates {
(0,2.29)+-(0,1.53)
(10,2.50)+-(0,1.16)
(20,4.07)+-(0,1.84)
(30,6.86)+-(0, 2.29)
(40,11.92)+-(0, 2.35)
		};

	\end{axis}
	\begin{axis} [name=plot3,at={($(plot1.south)-(0,0.1cm)$)},anchor=north,tick label style={font=\small},y label style={font=\small},ylabel=Test error ($\%$)]
	\node at (axis description cs:0.15,0.9) {covertype};

	\addplot+[error bars/.cd, y dir=both, y explicit, error mark=|,error mark options={scale=3.0}]
		coordinates {
			(0,23.21) +- (0,0.15)
			(10,23.22) +- (0,0.15)
			(20,25.12) +- (0,0.24)
			(30,29.95) +- (0,0.2)
			(40,40.02) +- (0,0.28)
		};

	\addplot+[error bars/.cd, y dir=both, y explicit, error mark=|,error mark options={scale=3.0}]
		coordinates {
			(0,23.6) +- (0,0.21)
			(10,23.36) +- (0,0.19)
			(20,25.43) +- (0,0.31)
			(30,31.6) +- (0,0.59)
			(40,42.38) +- (0,0.85)
		};

	\addplot+[error bars/.cd, y dir=both, y explicit, error mark=|,error mark options={scale=3.0}]
		coordinates {
			(0,23.85) +- (0,0.17)
			(10,23.27) +- (0,0.14)
			(20,25.27) +- (0,0.36)
			(30,30.63) +- (0,0.28)
			(40,40.4) +- (0,0.51)
		};

	\addplot+[error bars/.cd, y dir=both, y explicit, error mark=|,error mark options={scale=3.0}]
		coordinates {
			(0,23.02) +- (0,0.15)
			(10,23.36) +- (0,0.12)
			(20,25.12) +- (0,0.29)
			(30,32.94) +- (0,0.48)
			(40,42.31) +- (0,0.41)
		};
	\addplot+[error bars/.cd, y dir=both, y explicit, error mark=|,error mark options={scale=3.0}]
		coordinates {
(0,21.886400)+-(0,0.193000)
(10,22.082400)+-(0,0.158800)
(20,22.880500)+-(0,0.153500)
(30,25.220600)+-(0,0.319600)
(40,31.693400)+-(0,0.917000)
		};
	\addplot+[error bars/.cd, y dir=both, y explicit, error mark=|,error mark options={scale=3.0}]
		coordinates {
			(0,21.82) +- (0,.14)
			(10,21.98) +- (0,.13)
			(20,22.78) +- (0,.14)
			(30,24.67) +- (0,.29)
			(40,32.14) +- (0,.37)
		};
	\addplot+[error bars/.cd, y dir=both, y explicit, error mark=|,error mark options={scale=3.0}]
		coordinates {
			(0,22.31) +- (0,.12)
			(10,22.50) +- (0,0.12)
			(20,23.61) +- (0,.14)
			(30,27.12) +- (0,.33)
			(40,36.41) +- (0,.36)
		};
	\end{axis}
	\begin{axis} [name=plot4,at={($(plot2.south)-(0,0.1cm)$)},anchor=north,tick label style={font=\small}]
	\node at (axis description cs:0.19,0.9) {mushrooms};

	\addplot+[error bars/.cd, y dir=both, y explicit, error mark=|,error mark options={scale=3.0}]
		coordinates {
			(0,0.02) +- (0,0.06)
			(10,0.02) +- (0,0.06)
			(20,0.14) +- (0,0.27)
			(30,0.46) +- (0,0.55)
			(40,3.22) +- (0,1.14)
		};

	\addplot+[error bars/.cd, y dir=both, y explicit, error mark=|,error mark options={scale=3.0}]
		coordinates {
			(0,0.07) +- (0,0.12)
			(10,0.07) +- (0,0.12)
			(20,0.21) +- (0,0.34)
			(30,0.6) +- (0,0.66)
			(40,3.71) +- (0,0.89)
		};

	\addplot+[error bars/.cd, y dir=both, y explicit, error mark=|,error mark options={scale=3.0}]
		coordinates {
			(0,0.07) +- (0,0.12)
			(10,0.05) +- (0,0.09)
			(20,0.05) +- (0,0.12)
			(30,0.23) +- (0,0.48)
			(40,2.08) +- (0,0.84)
		};

	\addplot+[error bars/.cd, y dir=both, y explicit, error mark=|,error mark options={scale=3.0}]
		coordinates {
			(0,0.02) +- (0,0.06)
			(10,0.05) +- (0,0.12)
			(20,0.12) +- (0,0.19)
			(30,0.25) +- (0,0.37)
			(40,1.06) +- (0,1.41)
		};

\addplot+[error bars/.cd, y dir=both, y explicit, error mark=|,error mark options={scale=3.0}]
		coordinates {
(0,0.000000)+-(0,0.000000)
(10,0.052800)+-(0,0.112700)
(20,0.070400)+-(0,0.116800)
(30,0.088000)+-(0,0.118100)
(40,0.369700)+-(0,0.500800)
		};

	\addplot+[error bars/.cd, y dir=both, y explicit, error mark=|,error mark options={scale=3.0}]
		coordinates {
			(0,0.) +- (0,0.)
			(10,0.12) +- (0,0.16)
			(20,0.33) +- (0,0.095)
			(30,0.90) +- (0,0.24)
			(40,2.99) +- (0,.64)
		};
	\addplot+[error bars/.cd, y dir=both, y explicit, error mark=|,error mark options={scale=3.0}]
		coordinates {
			(0,0.) +- (0,0.)
			(10,0.) +- (0,0.)
			(20,0.) +- (0,0.)
			(30,0.) +- (0,0.)
			(40,0.29) +- (0,.33)
		};

	\end{axis}
	\begin{axis} [name=plot5,at={($(plot3.south)-(0,0.1cm)$)},anchor=north,x label style={font=\small},xlabel=Label noise ($\%$),xticklabels={0,10,20,30,40},y label style={font=\small},ylabel=Test error ($\%$),tick label style={font=\small}]
	\node at (axis description cs:0.13,0.9) {adult9};

	\addplot+[error bars/.cd, y dir=both, y explicit, error mark=|,error mark options={scale=3.0}]
		coordinates {
			(0,15.02) +- (0,0.4)
			(10,15.68) +- (0,0.53)
			(20,17.27) +- (0,0.29)
			(30,22.85) +- (0,0.61)
			(40,33.96) +- (0,0.39)
		};

	\addplot+[error bars/.cd, y dir=both, y explicit, error mark=|,error mark options={scale=3.0}]
		coordinates {
			(0,15.05) +- (0,0.39)
			(10,15.51) +- (0,0.57)
			(20,16.69) +- (0,0.67)
			(30,17.75) +- (0,0.4)
			(40,34.35) +- (0,0.48)
		};

	\addplot+[error bars/.cd, y dir=both, y explicit, error mark=|,error mark options={scale=3.0}]
		coordinates {
			(0,15.42) +- (0,0.49)
			(10,15.6) +- (0,0.51)
			(20,16.87) +- (0,0.49)
			(30,22.88) +- (0,0.6)
			(40,33.97) +- (0,0.39)
		};

	\addplot+[error bars/.cd, y dir=both, y explicit, error mark=|,error mark options={scale=3.0}]
		coordinates {
			(0,15.19) +- (0,0.56)
			(10,15.5) +- (0,0.43)
			(20,17.53) +- (0,0.47)
			(30,20.22) +- (0,0.94)
			(40,28.47) +- (0,0.54)
		};
	\addplot+[error bars/.cd, y dir=both, y explicit, error mark=|,error mark options={scale=3.0}]
		coordinates {
(0,14.945900)+-(0,0.423200)
(10,15.194500)+-(0,0.498600)
(20,16.200600)+-(0,0.496800)
(30,18.871000)+-(0,0.428400)
(40,25.791200)+-(0,0.593100)
		};
	\addplot+[error bars/.cd, y dir=both, y explicit, error mark=|,error mark options={scale=3.0}]
		coordinates {
			(0,14.82) +- (0,.45)
			(10,14.91) +- (0,.37)
			(20,15.78) +- (0,.42)
			(30,18.15) +- (0,.39)
			(40,25.60) +- (0,.65)
		};
	\addplot+[error bars/.cd, y dir=both, y explicit, error mark=|,error mark options={scale=3.0}]
		coordinates {
			(0,14.99) +- (0,.44)
			(10,15.07) +- (0,.43)
			(20,16.10) +- (0,.53)
			(30,17.65) +- (0,0.39)
			(40,24.91) +- (0,1.31)
		};

	\end{axis}
	\begin{axis} [name=plot6,at={($(plot4.south)-(0,0.1cm)$)},anchor=north,x label style={font=\small},xlabel=Label noise ($\%$),xticklabels={0,10,20,30,40},tick label style={font=\small}]
	\node at (axis description cs:0.12,0.9) {web8};

	\addplot+[error bars/.cd, y dir=both, y explicit, error mark=|,error mark options={scale=3.0}]
		coordinates {
			(0,1.14) +- (0,0.15)
			(10,1.31) +- (0,0.11)
			(20,1.38) +- (0,0.16)
			(30,1.64) +- (0,0.15)
			(40,3.81) +- (0,0.5)
		};

	\addplot+[error bars/.cd, y dir=both, y explicit, error mark=|,error mark options={scale=3.0}]
		coordinates {
			(0,1.5) +- (0,0.26)
			(10,1.47) +- (0,0.16)
			(20,1.55) +- (0,0.22)
			(30,1.95) +- (0,0.22)
			(40,2.42) +- (0,0.23)
		};

	\addplot+[error bars/.cd, y dir=both, y explicit, error mark=|,error mark options={scale=3.0}]
		coordinates {
			(0,1.33) +- (0,0.13)
			(10,1.34) +- (0,0.11)
			(20,1.42) +- (0,0.16)
			(30,1.7) +- (0,0.18)
			(40,2.41) +- (0,0.17)
		};

	\addplot+[error bars/.cd, y dir=both, y explicit, error mark=|,error mark options={scale=3.0}]
		coordinates {
			(0,1.17) +- (0,0.18)
			(10,1.43) +- (0,0.15)
			(20,1.45) +- (0,0.15)
			(30,1.63) +- (0,0.12)
			(40,1.99) +- (0,0.29)
		};

	\addplot+[error bars/.cd, y dir=both, y explicit, error mark=|,error mark options={scale=3.0}]
		coordinates {
(0,1.094800)+-(0,0.166800)
(10,1.181600)+-(0,0.105700)
(20,1.251500)+-(0,0.089200)
(30,1.519200)+-(0,0.249900)
(40,1.953200)+-(0,0.440700)
		};

	\addplot+[error bars/.cd, y dir=both, y explicit, error mark=|,error mark options={scale=3.0}]
		coordinates {
			(0,2.37) +- (0,0.21)
			(10,2.35) +- (0,0.25)
			(20,2.34) +- (0,0.23)
			(30,2.35) +- (0,0.26)
			(40,2.53) +- (0,0.30)
		};

	\addplot+[error bars/.cd, y dir=both, y explicit, error mark=|,error mark options={scale=3.0}]
		coordinates {
			(0,1.15) +- (0,0.10)
			(10,1.12) +- (0,0.10)
			(20,1.13) +- (0,0.12)
			(30,1.29) +- (0,0.11)
			(40,1.56) +- (0,0.075)
		};
	\end{axis}
\end{tikzpicture}
\vspace{.25cm}
 \caption{\label{error_vs_noise} Test error versus label noise for $7$ methods on $2$ synthetic data sets (Long-Servedio and Mease-Wyner) and $4$ real data sets from the UCI repository. Error bars are obtained from $10$-fold cross-validation with the hyperparameters recorded in the Appendix. As a stand-in for quantum annealing, a classical simulated annealing routine was used to optimize the sixth-order objective function and the 0-1 objective function. For each training cut at each selection of hyperparameters, we kept 50 classifiers having the lowest objective values of all states encountered. We computed validation error as the lowest of validation error produced by these 50 classifiers. This procedure is used for both 0-1 loss and sixth-order loss. Test error was obtained using the classifier of lowest validation error. This strategy is realistic as we expect that a quantum annealer will sample the lowest energy states as opposed to giving us only the global minima. q-loss was optimized using Tabu search in \cite{denchev2012}. We see that sixth-order loss outperforms the convex methods on every data set except for Long-Servedio and performs similarly to other non-convex methods on the other five sets.}
\end{figure*}

We find that for all real data sets, the sixth-order loss function outperforms all tested convex loss functions and performs similarly to the non-convex methods. The first two data sets shown in Figure \ref{error_vs_noise} are synthetic sets designed to break convex loss functions devised by Long and Servedio \cite{long2010} and Mease and Wyner \cite{mease2008}. The sixth-order loss performs poorly on the Long-Servedio set because the data set is designed so that the optimal solution has only extremely large margins and extremely small margins. The large margins dominate the risk minimization due to the steep walls of the sixth-order function and this forces all of the smaller margins very near zero where the sixth-order function is almost linear. We believe that the particular pathological behavior which leads to the poor performance of sixth-order loss on the Long-Servedio set is unlikely to occur in real data.

Figure 4 shows that the sixth-order loss function outperforms even the other non-convex methods on three of the four real data sets. However, we attribute the suboptimal q-loss and 0-1 loss results to a failure of the selected optimization routines rather than to a deficiency of the actual training objectives. One reason this seems likely is because sixth-order loss outperforms the other non-convex functions most significantly on web8 (the real data set with the greatest number of features) but losses to q-loss and 0-1 loss on covertype (the real data set with the fewest number of features). A comprehensive summary of the data sets is included in the Appendix. While non-convex, the sixth-order loss objective appears to be somewhat easier to optimize as a consequence of being significantly smoother than either the 0-1 loss or q-loss objective.

\section{Explicit tensor construction}

In this section we show how to represent any regularized risk objective using a polynomial loss function as PUBO. We first introduce a fixed bit-depth approximation. More substantially, we represent variables using a fixed-point representation as floating-point representations require a non-polynomial function of the bits. Using $d$ bits per feature, our encoding will require a total of $N = n d$ bits. The binary state vector $\bm{q} \in \mathbb{B}^{N}$ encodes the weight vector $\bm{w}$,
\begin{equation}
\bm{w}[i] \equiv \zeta \bm{q}[i d] - \zeta \sum_{j = 1}^{d-1} \left(\frac{1}{2}\right)^j \bm{q}[i d +j]
\end{equation}
where $\zeta \in \mathbb{R}$ determines the weight scale so that $\bm{w} \in \left(-\zeta, \zeta\right]^n$. Furthermore, we define a binary coefficient matrix, $\bm{\hat{k}} \in \mathbb{R}^{n \times N}$,
\begin{equation}
\bm{\hat{k}} \equiv \bm{I}^{n\times n} \otimes \left\langle \zeta, -\frac{\zeta}{2}, -\frac{\zeta}{4}, \hdots, \frac{\zeta}{2^{1-d}} \right\rangle
\end{equation}
where $\bm{I}^{n \times n}\otimes$ indicates a Kronecker product by an $n \times n$ identity matrix. This ``tiles'' the binary weight sequence into a stair-step pattern down the diagonal; e.g. if $n = 3$, $d = 2$ and $\zeta = 1$,
\begin{equation}
\bm{\hat{k}} = \left(\begin{matrix}
1 & -\frac{1}{2} & 0 & 0 & 0 & 0\\
0 & 0 & 1 & -\frac{1}{2} & 0 & 0\\
0 & 0 & 0 & 0 & 1 & -\frac{1}{2} \end{matrix}\right).
\end{equation}

\newpage

We do this because later on, it will be useful to think of $\bm{w}$ as a linear mapping of $\bm{q}$ given as $\bm{w} = \bm{\hat{k}} \bm{q}$. In general, any PUBO of degree $k$ can be expressed as,
\begin{equation}
E\left(\bm{q}\right) = \bm{v}^\top \bm{q}^{\otimes k}
\end{equation}
where $\bm{v} \in \mathbb{R}^{N^k}$ is a $k$-fold tensor and $\bm{q}^{\otimes k}$ represents the $k^\textrm{th}$ tensor power of $\bm{q}$. We now show how to obtain this embedding for a cubic loss function but do so in a way that is trivially extended to orders less than or greater than cubic. In terms of continuous weights the empirical risk objective may be expressed as,
\begin{align}
& f\left(\bm{w}\right) = \frac{1}{m}\sum_{i=0}^{m-1} \alpha_1 y_i \bm{x}_i^\top \bm{w} + \alpha_2 \left(\bm{x}_i^\top\bm{w}\right)^2 + \alpha_3 \left(y_i\bm{x}_i^\top \bm{w}\right)^3\nonumber\\
& =\underbrace{\left(\frac{\alpha_1}{m} \sum_{i=0}^{m-1} y_i \bm{x}_i^\top\right)}_{\bm{\varphi_1}^\top}\bm{w}
 + \underbrace{\left(\frac{\alpha_2}{m} \sum_{i=0}^{m-1} \left( \bm{x}_i^{\otimes 2}\right)^\top\right)}_{\bm{\varphi_2}^\top} \bm{w}^{\otimes 2}\nonumber\\
& + \underbrace{\left(\frac{\alpha_3}{m} \sum_{i=0}^{m-1} \left(y_i \bm{x}_i^{\otimes 3}\right)^\top\right)}_{\bm{\varphi_3}^\top}\bm{w}^{\otimes 3} = \sum_{j=1}^3 \bm{\varphi_j}^\top \bm{w}^{\otimes j} 
\end{align}
where
\begin{equation}
\bm{\varphi}_j = \frac{\alpha_j}{m} \sum_{i=0}^{m-1} \left(y_i \bm{x}_i\right)^{\otimes j}.
\end{equation}
Using tensor notation, $\ell_2$-norm regularization is
\begin{equation}
\Omega_2\left(\bm{w}\right) = \frac{\lambda_2}{2}\, \left(\bm{1}^{n^2}\right)^\top\! \bm{w}^{\otimes 2},
\end{equation}
where $\bm{1}^{n^2}$ denotes a vector of all ones with length $n^2$. We now use $\bm{\hat{k}}$ to expand the binary variable tensor,
\begin{eqnarray}
E \left(\bm{q}\right) & = &  f\left(\bm{w}\right) + \Omega_2\left(\bm{w}\right) \\
& = & \frac{\lambda_2}{2}\, \left(\bm{1}^{n^2}\right)^\top \!\left(\bm{\hat{k} q}\right)^{\otimes 2} + \,\sum_{j=1}^3 \bm{\varphi_j}^\top \left(\bm{\hat{k}} \bm{q}\right)^{\otimes j}.\nonumber
\end{eqnarray}
This expression implies the form of $\bm{v}$,
\begin{equation}
\bm{v} = \frac{\lambda_2}{2}\, \bm{1}^{n^2} \!\!\otimes \bm{\hat{k}}^{\otimes 2} +\sum_{j=1}^{3} \bm{\varphi_j} \otimes \bm{\hat{k}}^{\otimes j}.
\end{equation}
We slightly abuse notation in our definition of $\bm{v}$ by ``adding'' together tensors of different rank. To accomplish this the tensor of lower rank should be placed in a tensor having the same rank as the larger tensor by setting additional tensor indices equal to a lower tensor index. For instance, the element corresponding to $\left(i, j\right)$ in a tensor of rank two could be placed in a tensor of rank three at $\left(i, j, i\right)$ or $\left(i, j, j\right)$. We note that it is necessary to \emph{first} convert to binary and then combine the three terms; doing things the other way would introduce complications due to the fact that $w_i^r \neq w_i \,\, \forall i,r$ whereas $q_i^r = q_i\,\,\forall i,r$. Finally, we note that constructing $\bm{\varphi_3}^\top \otimes \hat{\bm{k}}^{\otimes 3}$ is the most computationally expensive part of this entire procedure taking $O\left(n^3 d^3 m\right)$ time.

This 3-fold tensor can be reduced to a QUBO matrix using ancillae. The optimal reduction is trivial using the tools developed in \cite{babbush2013}. In Appendix B of that paper, it shown that the number of ancillae which are required to collapse a fully connected cubic to 2-local is upper bounded by $\frac{N^2}{4}$. Again, the general bound for the quadratization of a PUBO of degree $k$ is  $O\left(N^{2 \log k}\right)$ \cite{boros2012}. This bound suggests that unlike prior encodings, the number of ancillae required is entirely independent of the number of training examples. We point out that the tensor form of this problem may be evaluated in a time that does not depend on the number of training examples.

\section{Conclusion}
We have introduced two unusual loss functions: the cubic loss function and the sixth-order loss function. Both losses are non-convex and show clear evidence of robustness to label noise. While superior to classically tractable convex training methods, both loss functions are highly parameterized and represent less than perfect approximations to 0-1 loss. Prima facie, this suggests that more popular non-convex loss functions, e.g. sigmoid loss, may be more reliable (or at least more straightforward) in some respects.

However, training under non-convex loss is formally NP-Hard and in order to obtain satisfactory solutions to such optimization problems, heuristic algorithms must query the objective function many times. Often, the quality of the eventual solution depends on the number of queries the heuristic is allowed. The fact that the polynomial loss functions may be compiled so that each query to the objective is independent of the number of training examples suggests that these loss functions may be more compatible with heuristic optimization routines. This same property ensures that these loss functions can be compiled to a Hamiltonian suitable for quantum annealing using a reasonable number of qubits (estimates of resources requirements for various example problems are included in the Appendix). This efficient embedding in quantum hardware makes binary classification under non-convex polynomial loss a promising target problem to accelerate using a quantum annealing machine.

\onecolumn
\section{Appendix}
\subsection{$\ell_0$-norm regularization}

While $\ell_0$ programming is well known to be NP-Hard, we believe that quantum annealing may allow us to obtain satisfactory minima in many instances. $\ell_0$-norm regularization is often used to train classifiers that are particularly efficient in terms of the number of features required for classification. For the situation in which we would like to train a classifier with binary weights, the regularization function is trivial,
\begin{equation}
\Omega_0\left(\bm{q}\right) = \lambda_0 \sum_{i=0}^{n-1} q_i.
\end{equation}
For multiple bit depth weights, $\ell_0$-norm regularization will require a modest number of ancilla qubits (one for each feature). Using our notation, the regularizer is
\begin{equation}
\Omega_0\left(\bm{q}\right) = \sum_{j=0}^{n-1} \left(\lambda_0 \, \bm{q}[N + j] + \phi \left(1-\bm{q}[N + j]\right)\sum_{k=0}^{d-1} \bm{q}[j d + k]\right).
\end{equation}
Minimizing $\Omega\left(\bm{q}\right)$ causes the ancillae $\bm{q}[N + j]$ to act as indicator bits, each of which is 1 if and only if $w_j \neq 0$ and is 0 otherwise. This is achieved by summing the binary variables that take part in a weight variable. Correctness comes from the that the binary representation of $w_j = 0$ is when all bits corresponding to that weight are 0. Thus, if even a single bit from weight $w_i$ is on, the objective will either incur a penalty of $\phi$ or will set the ancilla to 0 so as to obtain a penalty of $\lambda_0$ instead. Thus, as long as $\phi$ is sufficiently larger than $\lambda_0$, this function enforces $\ell_0$-norm regularization with weight of $\lambda_0$. This regularizer may be combined with the empirical risk function described previously.
\vspace{1cm}

\subsection{Sixth-order loss coefficients}
\begin{eqnarray}
\beta_0 & = & \frac{1}{3}+\frac{343 \left(125 \pi ^{3/2}-864 \,\Gamma \left(\frac{11}{6}\right)^3\right)}{1296 \left(343 \,\Gamma \left(\frac{11}{6}\right)^3+750 \,\Gamma \left(\frac{13}{6}\right)^3\right)-300125 \pi ^{3/2}}\\
\beta_1 & = & - \frac{35 \sqrt[6]{\omega } \left(245 \left(1+3\ 2^{2/3}\right) \pi  \,\Gamma \left(\frac{4}{3}\right)+36 \left(49 \,\Gamma \left(\frac{11}{6}\right) \left(\sqrt{\pi }-3 \,\Gamma \left(\frac{5}{3}\right) \,\Gamma \left(\frac{11}{6}\right)\right)-60 \,\Gamma \left(\frac{13}{6}\right)^2\right)\right)}{3 \left(-\frac{222950 \pi ^{3/2}}{9}+98784 \,\Gamma \left(\frac{11}{6}\right)^3+43200 \,\Gamma \left(\frac{13}{6}\right)^3\right)}\\
\beta_2 & = & \frac{2940 \sqrt{\pi } \sqrt[3]{\omega } \left(25 \sqrt{\pi } \,\Gamma \left(\frac{13}{6}\right)-42 \,\Gamma \left(\frac{11}{6}\right)^2\right)}{300125 \pi ^{3/2}-1296 \left(343 \,\Gamma \left(\frac{11}{6}\right)^3+750 \,\Gamma \left(\frac{13}{6}\right)^3\right)}\\
\beta_3 & = & \frac{3675 \sqrt{\pi } \sqrt{\omega } \left(7 \sqrt{\pi }+63 \left(\sqrt[3]{2}-1\right) \,\Gamma \left(\frac{5}{3}\right) \,\Gamma \left(\frac{11}{6}\right)+18 \left(2^{2/3}-3\right) \,\Gamma \left(\frac{4}{3}\right) \,\Gamma \left(\frac{13}{6}\right)\right)}{111475 \pi ^{3/2}-1296 \left(343 \,\Gamma \left(\frac{11}{6}\right)^3+150 \,\Gamma \left(\frac{13}{6}\right)^3\right)}\\
\beta_4 & = & \frac{2100 \sqrt{\pi } \omega ^{2/3} \left(49 \sqrt{\pi } \,\Gamma \left(\frac{11}{6}\right)-60 \,\Gamma \left(\frac{13}{6}\right)^2\right)}{432 \left(343 \,\Gamma \left(\frac{11}{6}\right)^3+750 \,\Gamma \left(\frac{13}{6}\right)^3\right)-\frac{300125 \pi ^{3/2}}{3}}\\
\beta_5 & = & \frac{7 \omega ^{5/6} \left(-7056 \,\Gamma \left(\frac{11}{6}\right)^2+1225 \left(3 \sqrt[3]{2}-1\right) \pi  \,\Gamma \left(\frac{5}{3}\right)+600 \,\Gamma \left(\frac{13}{6}\right) \left(7 \sqrt{\pi }-18 \,\Gamma \left(\frac{4}{3}\right) \,\Gamma \left(\frac{13}{6}\right)\right)\right)}{-\frac{222950 \pi ^{3/2}}{9}+98784 \,\Gamma \left(\frac{11}{6}\right)^3+43200 \,\Gamma \left(\frac{13}{6}\right)^3}
\end{eqnarray}

\clearpage
\subsection{Convergence of cubic loss function}

\begin{figure*}[!ht]
	\centering
	\begin{subfigure}{}
		\includegraphics[scale = 0.4]{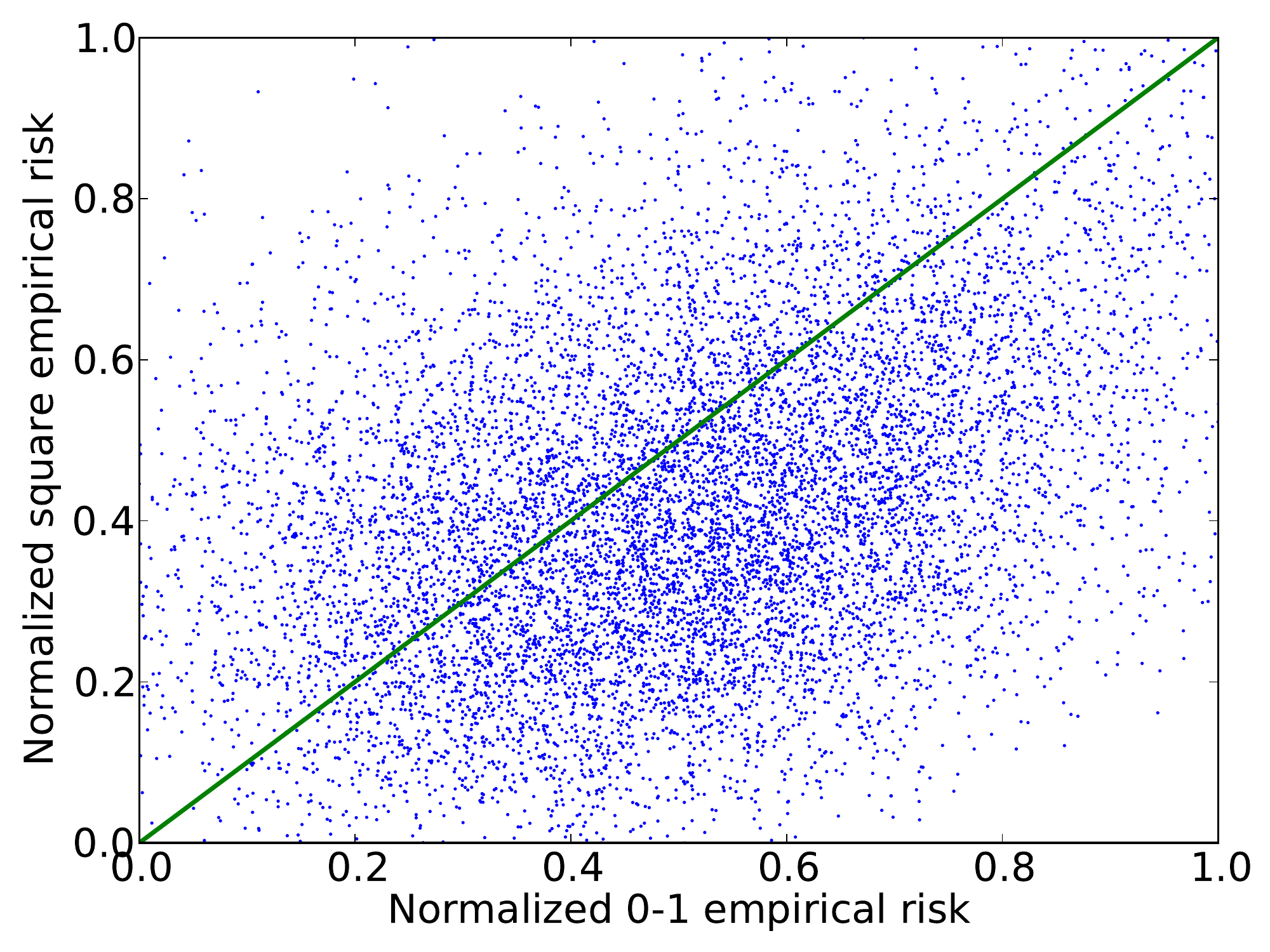}
	\end{subfigure}
	\begin{subfigure}{}
		\includegraphics[scale = 0.4]{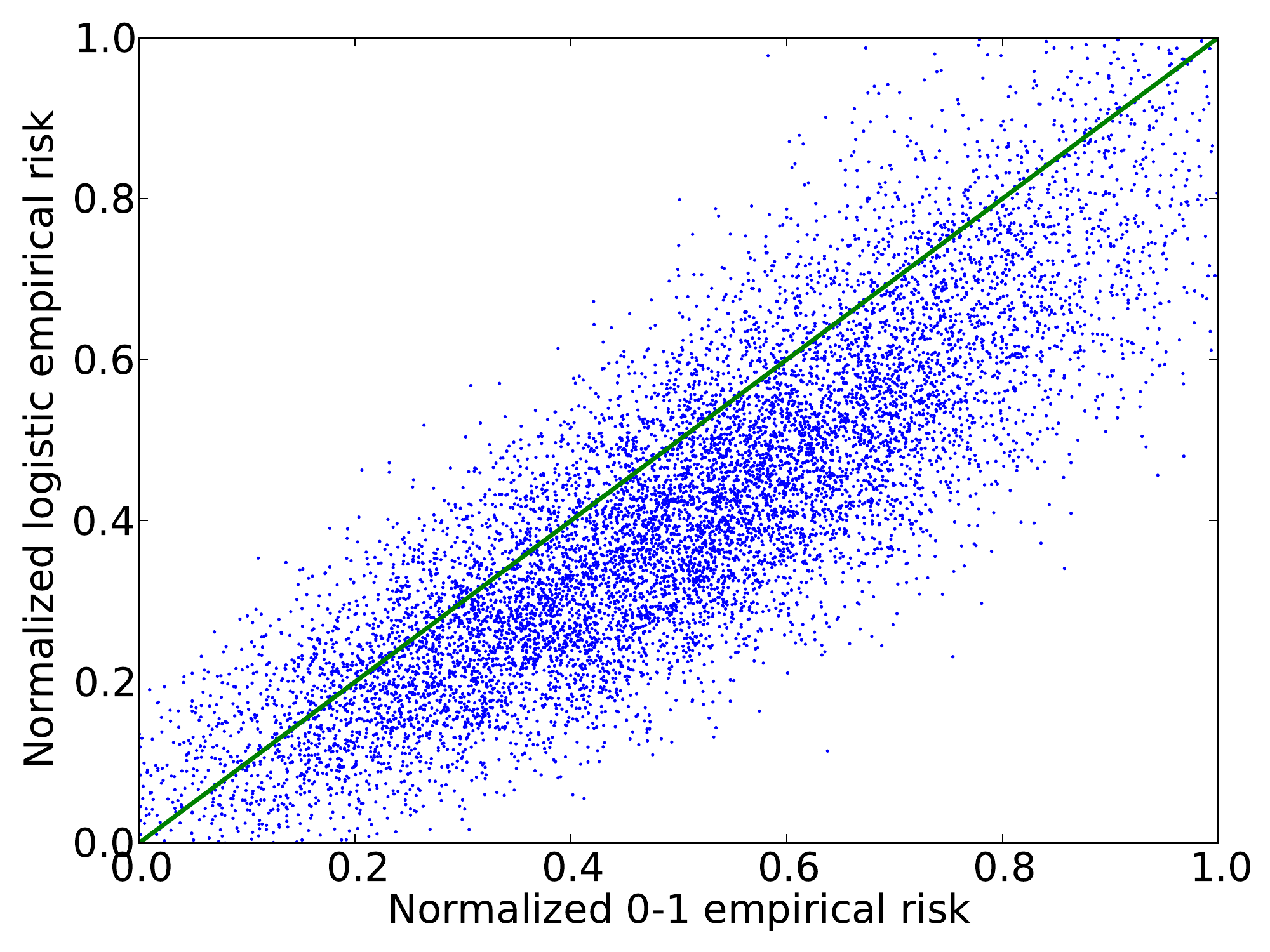}
	\end{subfigure}
	\begin{subfigure}{}
		\includegraphics[scale = 0.4]{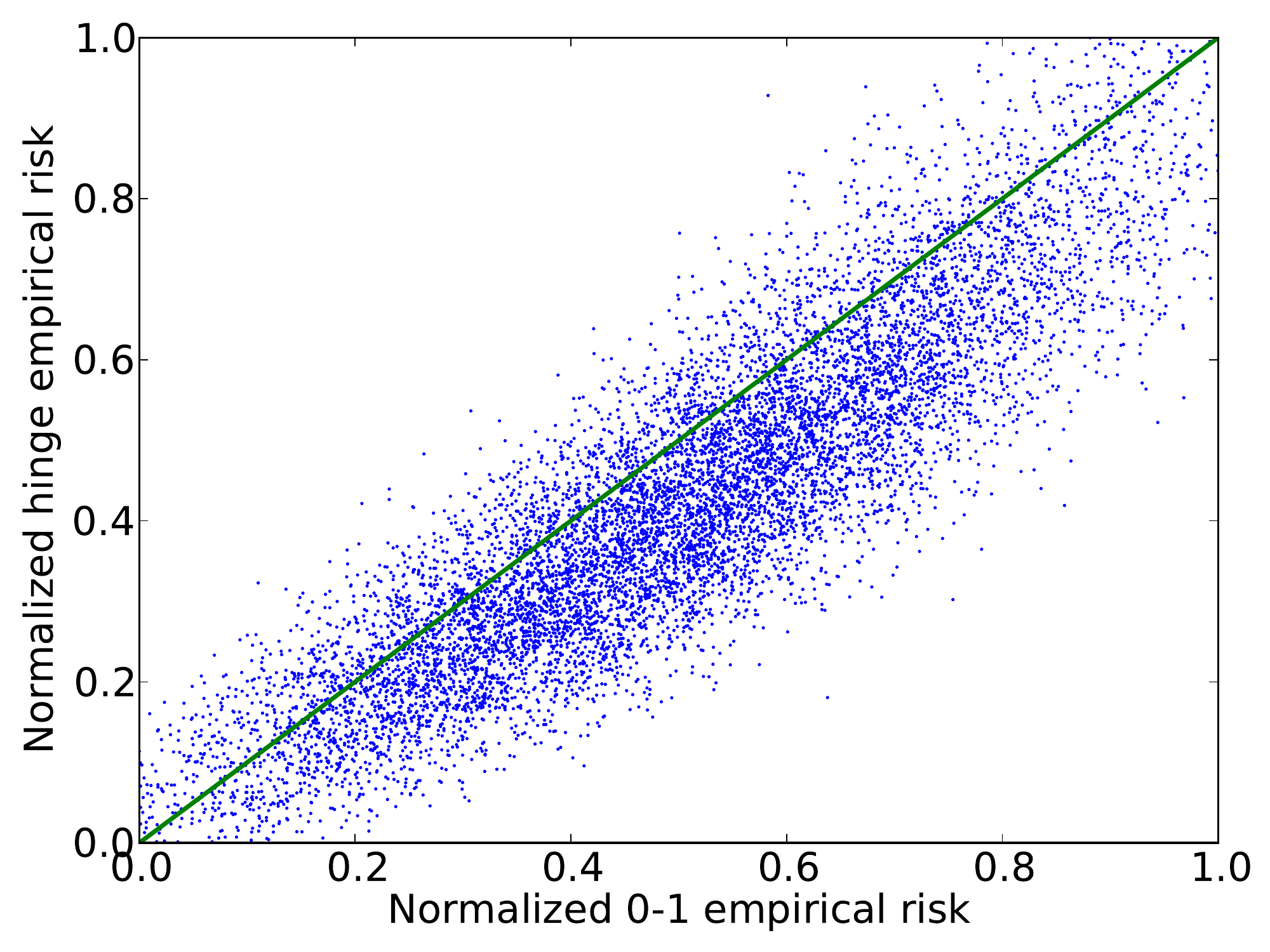}
	\end{subfigure}
	\begin{subfigure}{}
		\includegraphics[scale = 0.4]{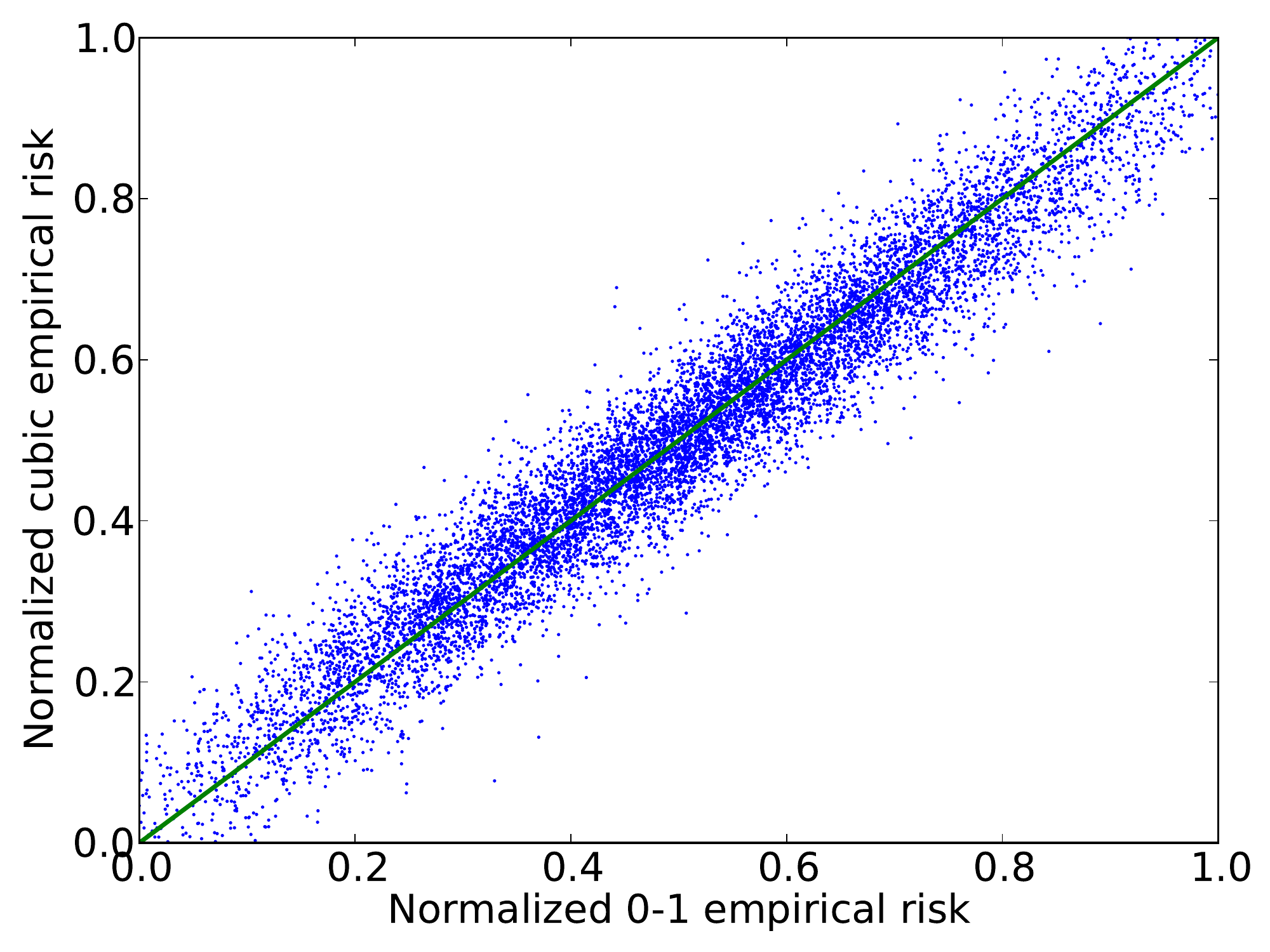}
	\end{subfigure}
	\begin{subfigure}{}
		\includegraphics[scale = 0.4]{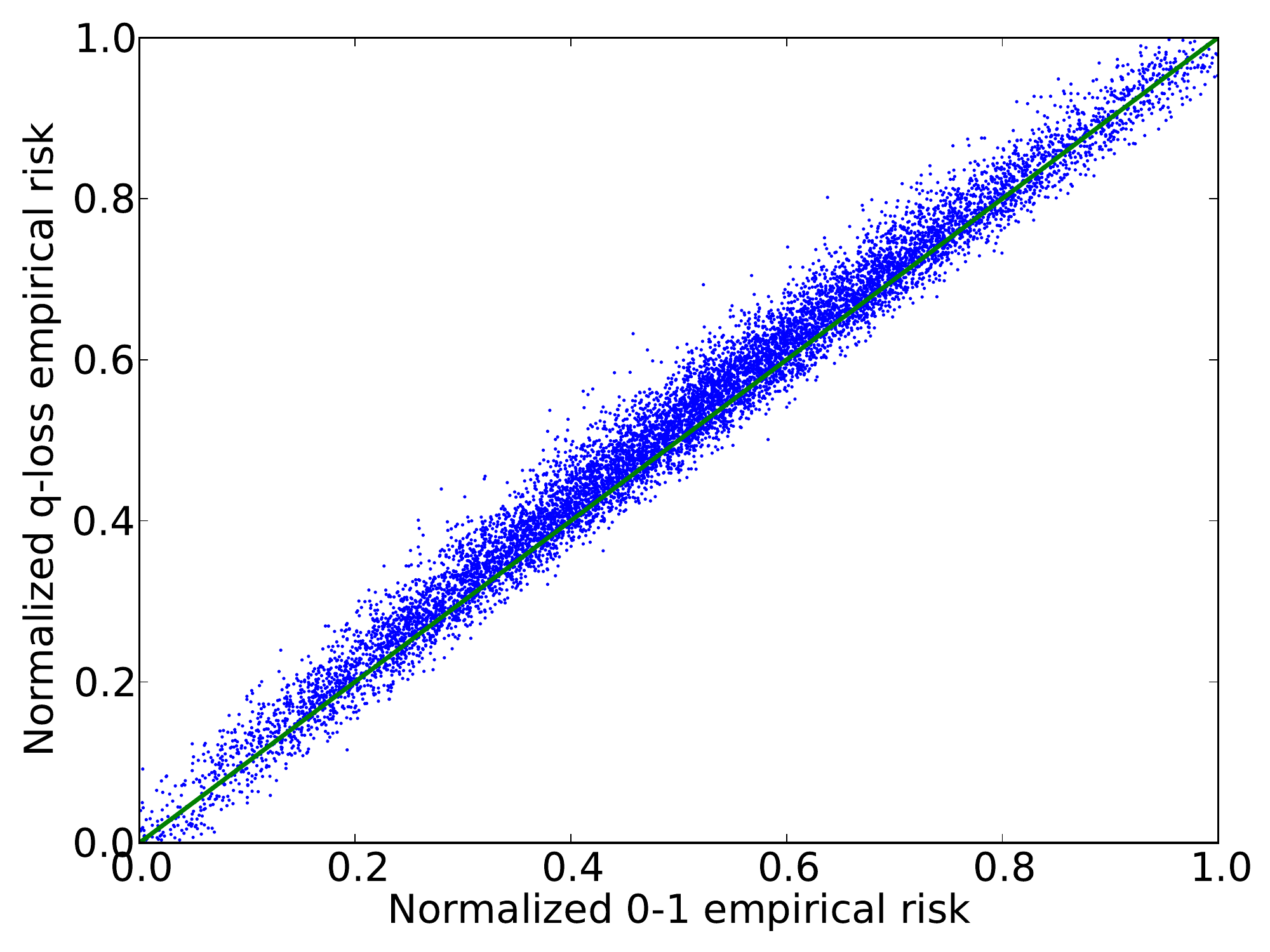}
	\end{subfigure}
	\begin{subfigure}{}
		\includegraphics[scale = 0.4]{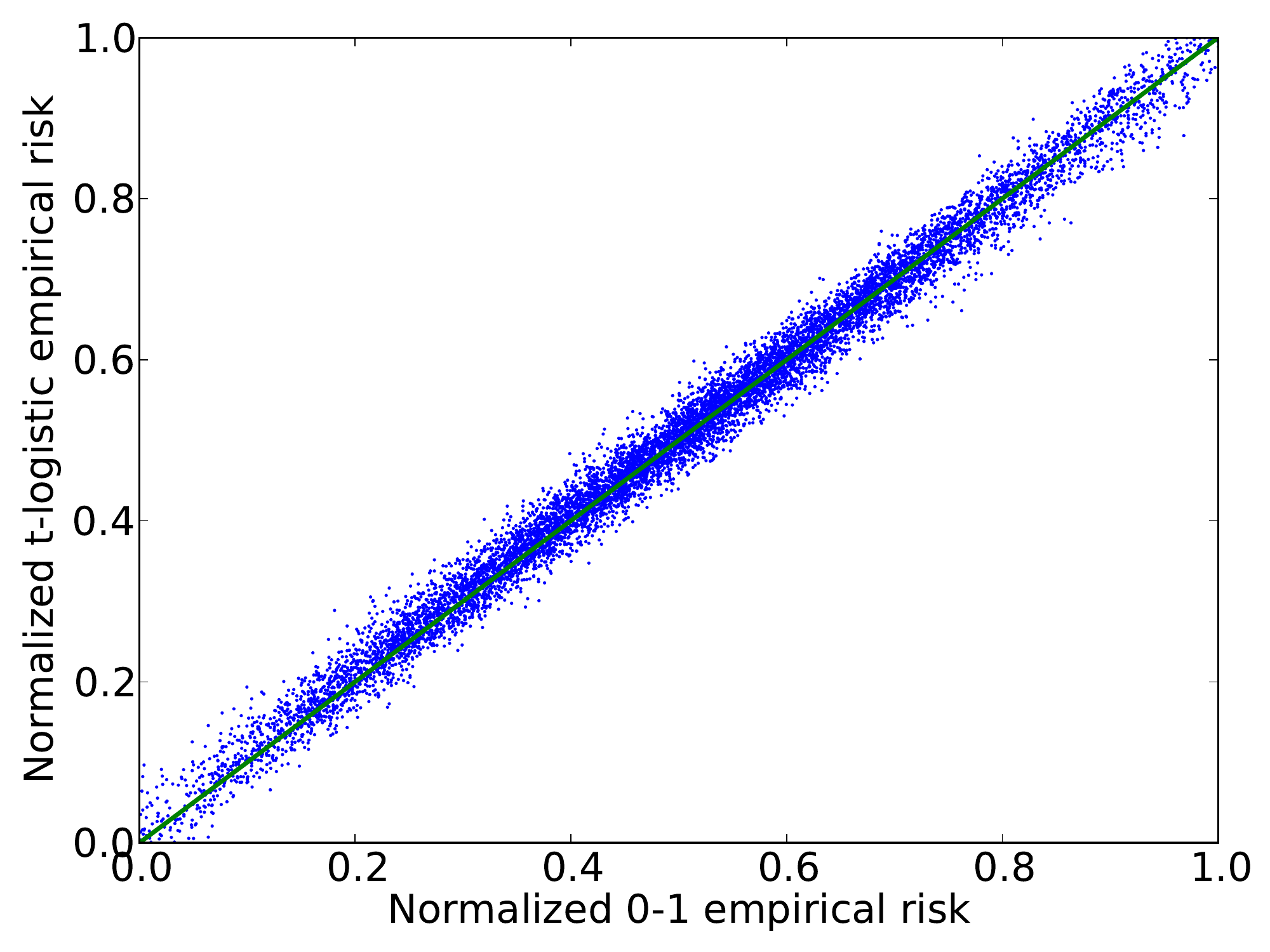}
	\end{subfigure}
  \caption{\label{correlations} Correlations between the total empirical risk of $10^4$ randomly selected states using various loss functions over $10^4$ training examples from the adult9 data set. The empirical risk values of each state have been uniformly shifted and rescaled to be in between 0 and 1. As we can see, the correlations between the convex loss functions and 0-1 loss are strictly worse than the correlation between cubic loss and 0-1 loss.}
\end{figure*}

\begin{figure*}[t]
	\centering
	\begin{subfigure}{}
		\includegraphics[scale = .4]{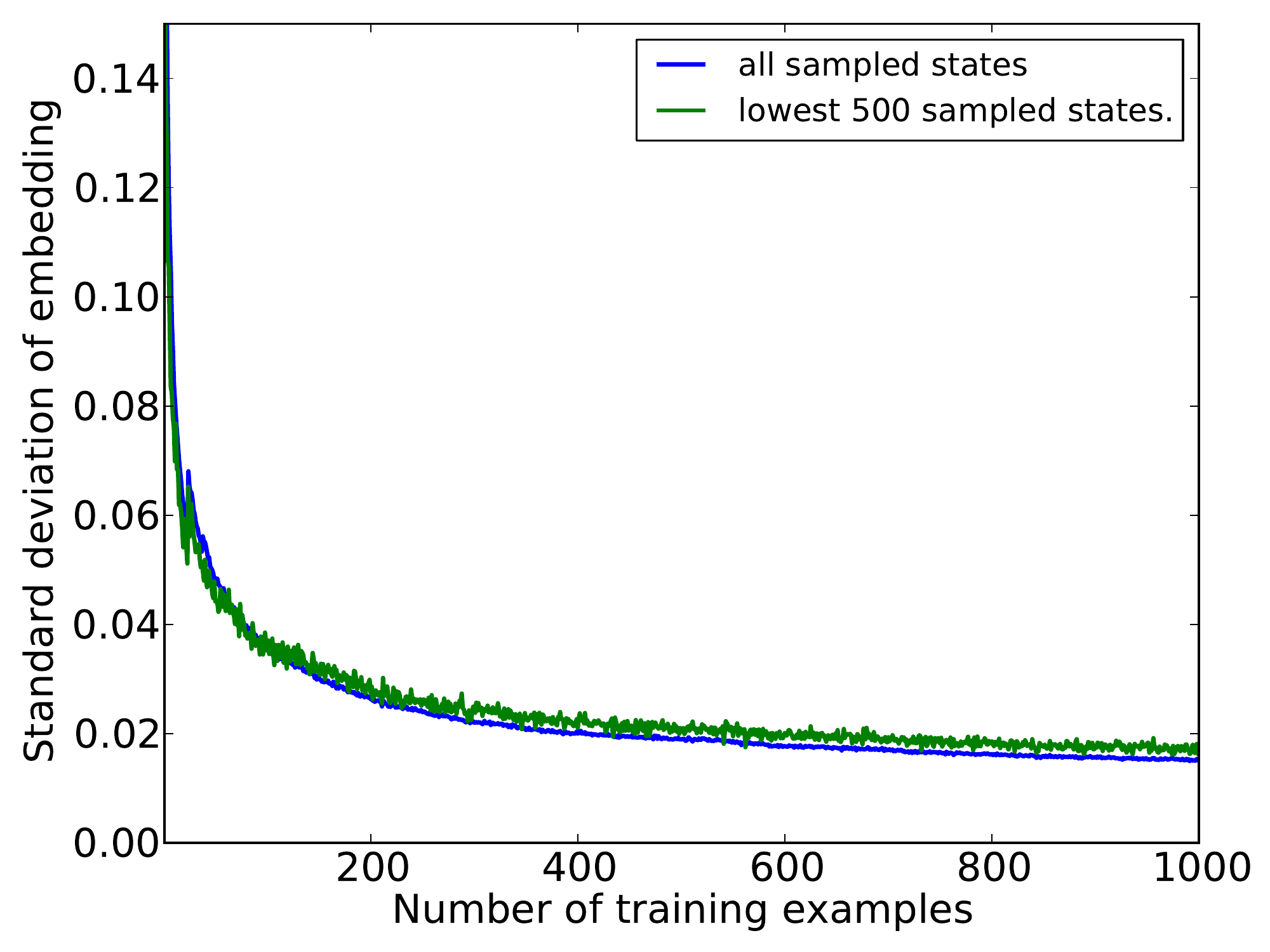}
	\end{subfigure}
		\begin{subfigure}{}
		\includegraphics[scale = .4]{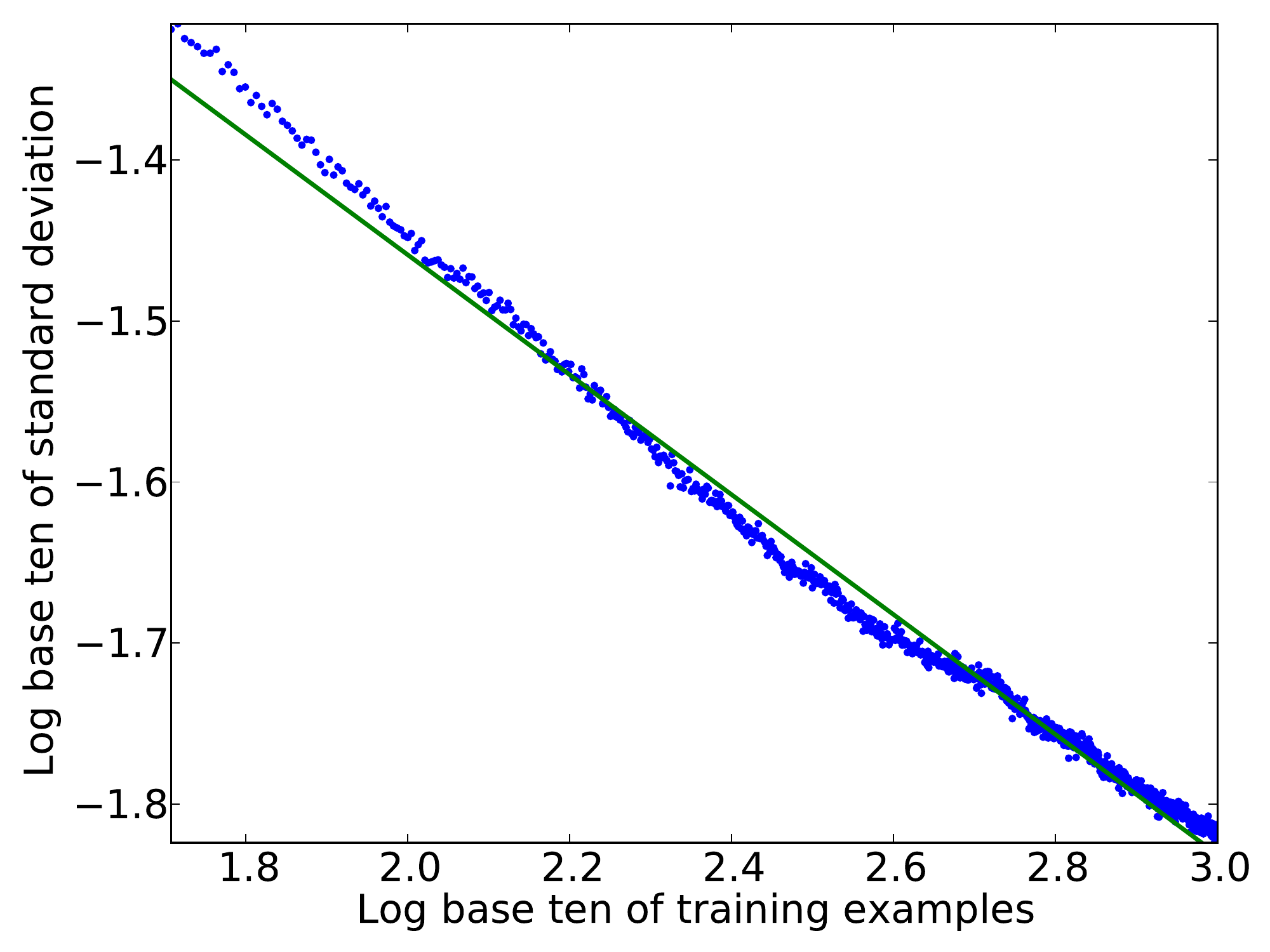}
	\end{subfigure}
	\begin{subfigure}{}
		\includegraphics[scale = 0.4]{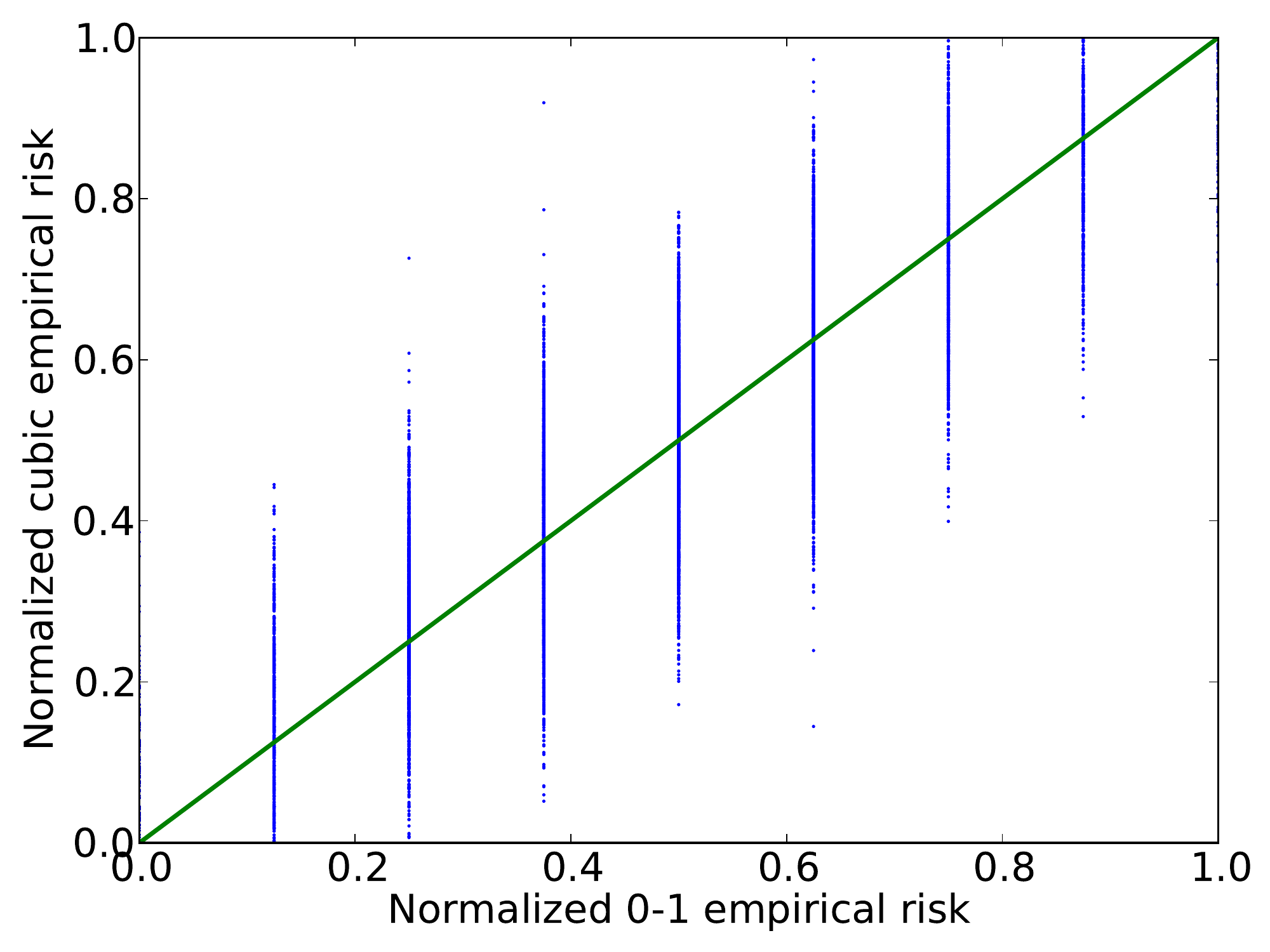}
	\end{subfigure}
	\begin{subfigure}{}
		\includegraphics[scale = 0.4]{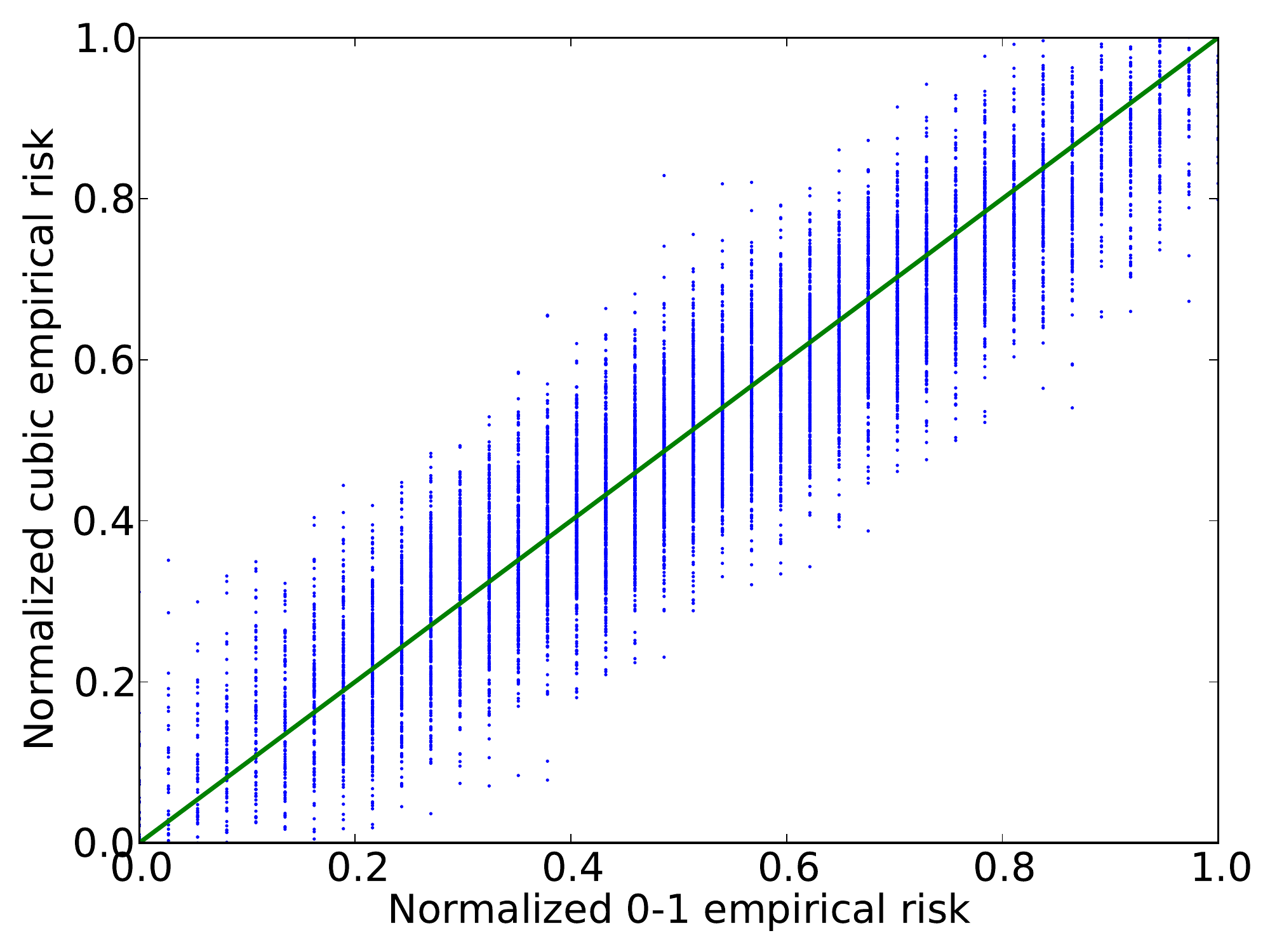}
	\end{subfigure}
	\begin{subfigure}{}
		\includegraphics[scale = 0.4]{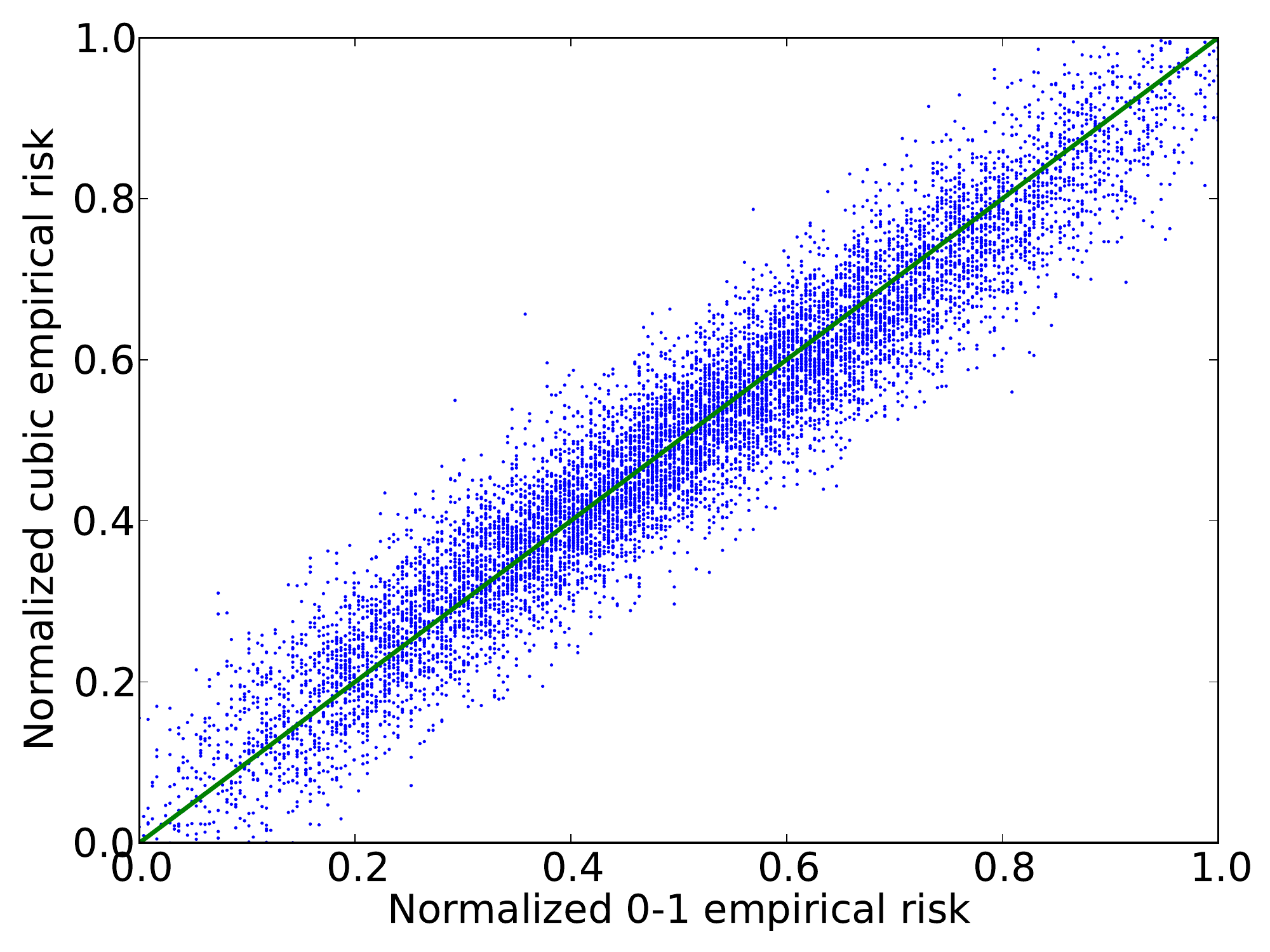}
	\end{subfigure}
	\begin{subfigure}{}
		\includegraphics[scale = 0.4]{./correlations_10000.pdf}
	\end{subfigure}
  \caption{These plots quantify the convergence of the empirical risk embedding error as a function of the number of training examples. The upper-left plot is made by fitting the cubic loss function and evaluating the resultant embedding error on adult9 using a variable number of training examples. Here, the error is the standard deviation of the energy landscapes defined by the limited number of training examples. At each point, $10^4$ states are sampled at random to evaluate the error. We also show the error in only the lowest 50 of these states to give an indication of the rate at which the low energy subspace is converging. On the upper-right, is a log plot of the same data indicating that embedding appears to converge as roughly $O\left(m^{-1/3}\right)$. The remaining plots show correlations between the total empirical risk of the $10^4$ randomly selected states using 0-1 loss and the total empirical risk on those states using cubic loss. These four plots were obtained by fitting the cubic loss function to the adult9 data set using: 10 training examples, 100 training examples, 1,000 training examples and 10,000 training examples. The error in these embeddings are 0.107, 0.0818, 0.062 and 0.055, respectively.}
\end{figure*}
\clearpage

\subsection{Estimated qubit requirements}
\begin{table*}[h]
\caption{Upper-bounds on qubit requirements for selected problems}
\centering
\begin{tabular}{ccccc}
\toprule
Loss function degree & \#Features & Bit-depth & \#Qubits\\
\midrule
\rowcolor[gray]{0.9}
cubic & 100 & 1 & 2,550\\
cubic & 100 & 4 & 40,200\\
\rowcolor[gray]{0.9}
cubic & 500 & 1 & 62,750\\
cubic & 500 & 4 & 1,001,000\\
\rowcolor[gray]{0.9}
cubic & 2,500 & 1 & 1,563,750\\
cubic & 2,500 & 4 & 25,005,000\\
\bottomrule
\end{tabular}\label{table:data_summary}
\end{table*}
\vspace{1.25cm}

\subsection{Data summary}
\begin{table*}[h]
\caption{Data summary}
\centering
\begin{tabular}{cccccc}
\toprule
Name & Dims & \#Examples & Density ($\%$) & Baseline error ($\%$)\\
\midrule
\rowcolor[gray]{0.9}
Long-Servedio & 21 & 2000 & 100.00 & 50.00\\
Mease-Wyner & 20 & 2000 & 100.00 & 49.80\\
\rowcolor[gray]{0.9}
covertype & 54 & 581012 & 22.20 & 36.46\\
mushrooms & 112 & 8124 & 18.75 & 48.20\\
\rowcolor[gray]{0.9}
adult9 & 123 & 48842 & 11.30 & 23.93\\
web8 & 300 & 59245 & 4.20 & 2.92\\
\bottomrule
\end{tabular}\label{table:data_summary}
\end{table*}
\vspace{1.25cm}

\subsection{Hyperparameters}
\begin{table*}[h]
\begin{minipage}[b]{0.3\linewidth}
\caption {values of $\lambda$ and $\omega$ offered to cross-validation}
\centering
\begin{tabular}{c}
\toprule
$\lambda$ \& $\omega$ \\
\midrule
\rowcolor[gray]{0.9}
2.000000\\
0.398965\\
\rowcolor[gray]{0.9}
0.079583\\
0.015875\\
\rowcolor[gray]{0.9}
0.003167\\
0.000632\\
\rowcolor[gray]{0.9}
0.000126\\
0.000025\\
\rowcolor[gray]{0.9}
0.000005\\
0.000001\\
\bottomrule
\end{tabular}\label{table:lambda_values}
\end{minipage}
\begin{minipage}[b]{0.75\linewidth}
\caption {$\omega$ values for sixth-order loss picked by cross-validation}
\centering
\begin{tabular}{cccccc}
\toprule
\multirow{2}{*}{Data set name} & \multicolumn{5}{c}{Label noise ($\%$)} \\
\cmidrule(r){2-6}
& 0 & 10 & 20 & 30 & 40 \\
\midrule
\rowcolor[gray]{0.9}
Long-Servedio  & 0.000001 & 0.000001 & 0.000001 & 0.000001 & 0.000632\\
Mease-Wyner & 0.398965 & 0.000126 & 0.000126 & 0.000025 & 0.000005\\
\rowcolor[gray]{0.9}
covertype & 2.000000 & 2.000000 & 0.398965 & 2.000000 & 0.000025\\
mushrooms & 2.000000 & 2.000000 & 2.000000 & 0.000632 & 0.000005\\
\rowcolor[gray]{0.9}
adult9 & 0.000001 & 2.000000 & 0.000025 & 0.079583 & 2.000000\\
web8 & 0.000632 & 0.000126 & 0.398965 &  0.398965 & 2.000000 \\
\bottomrule
\end{tabular}\label{table:c_liblinear}
\end{minipage}
\end{table*}

\begin{table}[h]
\caption {$q$ values for $q$-loss picked by cross-validation}
\centering
\begin{tabular}{cccccc}
\toprule
\multirow{2}{*}{Data set name} & \multicolumn{5}{c}{Label noise ($\%$)} \\
\cmidrule(r){2-6}
& 0 & 10 & 20 & 30 & 40 \\
\midrule
\rowcolor[gray]{0.9}
Long-Servedio  & 0 & -0.39 & -0.24 & -0.71 & -0.55\\
Mease-Wyner & 0 & -2.96 & -1.62 & -1.36 & 0\\
\rowcolor[gray]{0.9}
covertype & -0.63 & -0.54 & -0.38 & -0.5 & -0.51\\
mushrooms & 0 & -0.76 & -0.47 & -0.17 & -0.13\\
\rowcolor[gray]{0.9}
adult9 & -0.86 & -0.53 & -0.43 & -0.53 & -0.07\\
web8 & -0.99 & -0.46 & -0.41 & -0.19 & 0\\
\bottomrule
\end{tabular}\label{table:q_value}
\end{table}

\begin{table*}[h]
\caption {$C$ values for liblinear picked by cross-validation}
\centering
\begin{tabular}{cccccc}
\toprule
\multirow{2}{*}{Data set name} & \multicolumn{5}{c}{Label noise ($\%$)} \\
\cmidrule(r){2-6}
& 0 & 10 & 20 & 30 & 40 \\
\midrule
\rowcolor[gray]{0.9}
Long-Servedio  & 0.499978 & 2.506486 & 0.499978 & 0.499978 & 0.499978\\
Mease-Wyner & 40000.00 & 0.499978 & 315.7562 & 12.565498 & 62.99213\\
\rowcolor[gray]{0.9}
covertype & 0.499978 & 2.506486 & 62.99213 & 1000000.0 & 12.56541\\
mushrooms & 2.506486 & 12.56541 & 0.499978 & 0.499978 & 0.499978\\
\rowcolor[gray]{0.9}
adult9 & 0.499978 & 62.992126 & 0.499978 & 0.499978 & 0.499978\\
web8 & 315.7562 & 0.499978 & 12.56541 & 12.56541 & 0.499978\\
\bottomrule
\end{tabular}\label{table:c_liblinear}
\end{table*}

\begin{table*}[h]
\caption {$\lambda$ values picked by cross-validation for $0\%$ label noise}
\centering
\begin{tabular}{cccccc}
\toprule
\multirow{2}{*}{Data set name} & \multicolumn{5}{c}{Method} \\
\cmidrule(r){2-6}
& logistic & square & smooth hinge & q-loss & sixth-order\\
\midrule
\rowcolor[gray]{0.9}
Long-Servedio & 0.003167 & 0.079583 & 0.015875 & 0.015875 & 0.000001\\
Mease-Wyner & 0.000001 & 0.000025 & 0.000001 & 0.000126 & 2.000000\\
\rowcolor[gray]{0.9}
covertype & 0.000025 & 0.000025 & 0.000001 & 0.000025 & 0.000632\\
mushrooms & 0.000001 & 0.000025 & 0.000632 & 0.000025 & 0.398965\\
\rowcolor[gray]{0.9}
adult9 & 0.000001 & 0.000632 & 0.000126 & 0.003167 & 0.015875\\
web8 & 0.000001 & 0.000005 & 0.000001 & 0.000632  & 0.015875\\
\bottomrule
\end{tabular}\label{table:lambda_noise0}
\end{table*}

\begin{table*}[h]
\caption {$\lambda$ values picked by cross-validation for $10\%$ label noise}
\centering
\begin{tabular}{cccccc}
\toprule
\multirow{2}{*}{Data set name} & \multicolumn{5}{c}{Method} \\
\cmidrule(r){2-6}
& logistic & square & smooth hinge & q-loss & sixth-order\\
\midrule
\rowcolor[gray]{0.9}
Long-Servedio & 0.000005 & 2.000000 & 0.003167 & 0.015875  & 0.000001\\
Mease-Wyner & 0.000005 & 0.000632 & 0.000005 & 0.000126 & 0.398965\\
\rowcolor[gray]{0.9}
covertype & 0.000025 & 0.000632 & 0.000126 & 0.000001 & 0.000001\\
mushrooms & 0.000005 & 0.000001 & 0.000005 & 0.003167 & 0.398965\\
\rowcolor[gray]{0.9}
adult9 & 0.000632 & 0.003167 & 0.000126 & 0.015875 & 0.000632\\
web8 & 0.000005 & 0.000126 & 0.000005 & 0.000632 & 0.015875\\
\bottomrule
\end{tabular}\label{table:lambda_noise10}
\end{table*}
\FloatBarrier

\makeatletter
\setlength{\@fptop}{5pt}
\makeatother
\begin{table*}[h]
\caption {$\lambda$ values picked by cross-validation for $20\%$ label noise}
\centering
\begin{tabular}{cccccc}
\toprule
\multirow{2}{*}{Data set name} & \multicolumn{5}{c}{Method} \\
\cmidrule(r){2-6}
& logistic & square & smooth hinge & q-loss & sixth-order\\
\midrule
\rowcolor[gray]{0.9}
Long-Servedio & 2.000000 & 2.000000 & 2.000000  & 0.000126 & 0.000001\\
Mease-Wyner & 0.000025 & 0.000005 & 0.000126 & 0.000126 & 0.079583\\
\rowcolor[gray]{0.9}
covertype & 0.000001 & 0.000126 & 0.000126 & 0.000001 & 0.000025\\
mushrooms & 0.000126 & 0.000632 & 0.000025 & 0.003167 & 0.398965\\
\rowcolor[gray]{0.9}
adult9 & 0.079583 & 0.079583 & 0.003167  & 0.015875 & 2.000000\\
web8 & 0.000001 & 0.000001 & 0.000126 & 0.000632 & 0.015875\\
\bottomrule
\end{tabular}\label{table:lambda_noise20}
\end{table*}

\begin{table*}[ht]
\caption {$\lambda$ values picked by cross-validation for $30\%$ label noise}
\centering
\begin{tabular}{cccccc}
\toprule
\multirow{2}{*}{Data set name} & \multicolumn{5}{c}{Method} \\
\cmidrule(r){2-6}
& logistic & square & smooth hinge & q-loss & sixth-order\\
\midrule
\rowcolor[gray]{0.9}
Long-Servedio & 2.000000 & 2.000000 & 2.000000 & 0.003167 & 0.000001\\
Mease-Wyner & 0.000005 & 0.000001 & 0.000005  & 0.000126 & 0.000632\\
\rowcolor[gray]{0.9}
covertype & 0.000001 & 0.000126 & 0.000025 & 0.000025  & 0.398965\\
mushrooms & 0.000632 & 0.003167 & 0.000632 & 0.003167 & 0.079583\\
\rowcolor[gray]{0.9}
adult9 & 2.000000 & 0.003167  & 2.000000 & 0.003167& 0.000632\\
web8 & 0.000126 & 0.000001 & 0.000126 & 0.000632  & 0.000005\\
\bottomrule
\end{tabular}\label{table:lambda_noise30}

\vspace{1cm}
\caption {$\lambda$ values picked by cross-validation for $40\%$ label noise}
\centering
\begin{tabular}{cccccc}
\toprule
\multirow{2}{*}{Data set name} & \multicolumn{5}{c}{Method} \\
\cmidrule(r){2-6}
& logistic & square & smooth hinge & q-loss & sixth-order\\
\midrule
\rowcolor[gray]{0.9}
Long-Servedio & 2.000000 & 2.000000 & 2.000000 & 0.003167 & 0.000632\\
Mease-Wyner & 0.000001 & 0.000005  & 0.000025  & 0.000126 & 0.000632\\
\rowcolor[gray]{0.9}
covertype & 0.000001 & 0.000001 & 0.000001 & 0.000001 & 0.079583\\
mushrooms & 0.000126 & 0.000632 & 0.003167 & 0.003167 & 0.003167\\
\rowcolor[gray]{0.9}
adult9 & 0.000126 & 0.000126 & 0.079583  & 0.000025 & 0.000025\\
web8 & 0.015875 & 0.079583 & 0.000632 & 0.000632 & 0.000001\\
\bottomrule
\end{tabular}\label{table:lambda_noise40}
\end{table*}

\begingroup
\let\clearpage\relax
\bibliographystyle{icml2012}
\bibliography{library}
\endgroup

\end{document}